\documentclass[10pt,twocolumn,letterpaper]{article}

\usepackage[pagenumbers]{cvpr} 

\usepackage{graphicx}
\usepackage{amsmath}
\usepackage{amssymb}
\usepackage{booktabs}
\usepackage{makecell} 
\usepackage{array} 
\usepackage{bbm}  

\newcommand{\afterfigure}{\vspace{-1em}}
\newcommand\norm[1]{\left\lVert#1\right\rVert}
\newcommand{\vv}[1]{\mathbf{#1}}

\usepackage[pagebackref,breaklinks,colorlinks]{hyperref}

\usepackage[capitalize]{cleveref}
\crefname{section}{Sec.}{Secs.}
\Crefname{section}{Section}{Sections}
\Crefname{table}{Table}{Tables}
\crefname{table}{Tab.}{Tabs.}

\def\confName{CVPR}
\def\confYear{2023}

\begin{document}

\title{Neural Congealing: Aligning Images to a Joint Semantic Atlas}

\author{Dolev Ofri-Amar$^1$ \qquad
Michal Geyer$^1$ \qquad
Yoni Kasten$^2$ \qquad
Tali Dekel$^1$ \vspace{0.2cm}\\ 
\vspace{0.2cm}
{\normalsize $^1$Weizmann Institute of Science \qquad $^2$NVIDIA Research} \\
{\small Project webpage: \url{https://neural-congealing.github.io/}} 
}

\twocolumn[{
\renewcommand\twocolumn[1][]{#1}
\maketitle
\centering
\vspace*{-0.55cm}
\includegraphics[width=\textwidth]{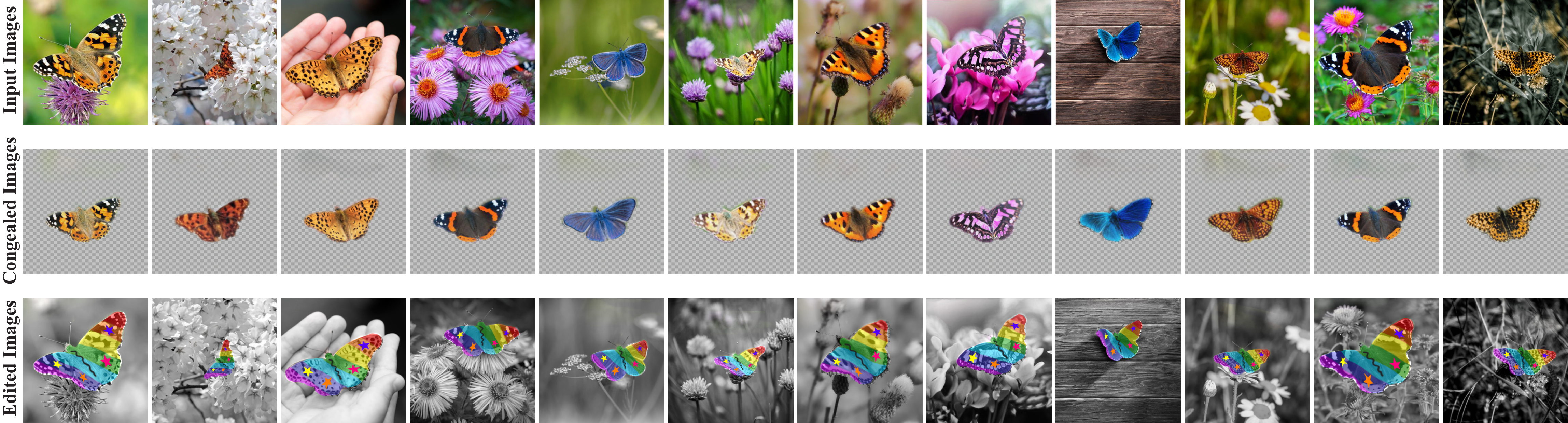}
\captionof{figure}{Given a set of input images, our method automatically detects and jointly aligns semantically-common content across the images. This is achieved through a test-time training approach that estimates a unified 2D  atlas  that represents the common semantic content, and dense mappings from the joint atlas to each of the input images. Our atlas and mappings are optimized per input set in a self-supervised manner by leveraging a pre-trained DINO-ViT model. Our method can be applied to diverse image sets, without requiring any additional training data, and allows us to automatically propagate an edit applied to a single image across the entire set.
}
\label{fig:teaser}
 \vspace*{0.65cm}
}]

\begin{abstract}\vspace{-0.2cm}
We present Neural Congealing -- a zero-shot self-supervised framework for detecting and jointly aligning semantically-common content across a given set of  images.  Our approach harnesses the power of pre-trained DINO-ViT features to learn: (i) a joint semantic atlas -- a 2D grid that captures the mode of DINO-ViT features in the input set, and (ii) dense mappings from the unified atlas to each of the input images. We derive a new robust self-supervised framework that optimizes the atlas representation and mappings per image set, requiring  only a few real-world images as input without any additional input information (e.g., segmentation masks). Notably, we design our losses and training paradigm to account only for the shared content under severe variations in appearance, pose,  background clutter or other distracting objects.  We demonstrate results on a plethora of challenging image sets   including sets of mixed domains (e.g., aligning images depicting sculpture and artwork of cats), sets depicting related yet different object categories (e.g., dogs and tigers), or domains for which large-scale training data is scarce (e.g., coffee mugs). We thoroughly evaluate our method and show that our test-time optimization approach performs favorably compared to a state-of-the-art method that requires extensive training on large-scale datasets. 

\end{abstract}

\vspace{-5mm}
\section{Introduction}
\label{sec:intro}

Humans can easily associate and match semantically-related objects across images, even under severe variations in appearance, pose and background content. For example, by observing the images in Fig.~\ref{fig:teaser}, we can immediately focus and visually compare the different butterflies, while ignoring the rest of the irrelevant content. While computational methods for establishing \emph{semantic} correspondences have seen a significant  progress in recent years, research efforts are largely focused on either estimating \emph{sparse} matching across \emph{multiple} images (e.g., keypoint detection), or establishing \emph{dense} correspondences between a \emph{pair of images}. In this paper, we consider the task of joint \emph{dense semantic alignment of multiple images}. Solving this long-standing task is useful for a variety of applications, ranging from editing image collections \cite{zhu2014averageexplorer,peebles2022gan}, browsing images through canonical primitives, and 3D reconstruction 
 (e.g., \cite{Tang_2021_ICCV,facciolo2017automatic}).

The task of joint image alignment dates back to the seminal congealing~\cite{learned2005data,huang2007unsupervised,miller2000learning,huang2012learning}, which aligns a set of images into a common 2D space. Recently,  GANgealing \cite{peebles2022gan} has modernized this approach for congealing an entire \emph{domain} of images. This is achieved by leveraging a pre-trained GAN to generate images that serve as self-supervisory signal. Specifically, their method jointly learns both the mode of the generated images in the latent space of the GAN, and a network that predicts the mappings of the images into the joint mode.  GANgealing demonstrated impressive results on in-the-wild image sets. 
Nevertheless, their method requires a StyleGAN model pre-trained on the domain of the test images, e.g., aligning cat images requires  training  StyleGAN on a large-scale cat dataset. This is a challenging task by itself, especially for unstructured image domains or uncurated datasets~\cite{mokady2022self}. Furthermore, they require additional extensive training for learning the mode of the generated images and their mapping network (e.g., training on millions of generated images).   
In this work, we take a different route and tackle the joint alignment task in the challenging setting where only a test image set is available, without any additional training data. More specifically,  given only a few images as input (e.g., $<$25 images), our method estimates the mode of the test set and their joint dense alignment, in a self-supervised manner. 

We assume that the input images share a common semantic content, yet may depict various factors of variations, such as pose, appearance, background content or other distracting objects (e.g., \emph{Mugs} in Fig.~\ref{fig:results}). We take inspiration from the tremendous progress in representation learning, and leverage a pre-trained DINO-ViT -- a Vision Transformer model trained in a  self-supervised manner~\cite{caron2021emerging}. DINO-ViT features have been shown to serve as an effective visual descriptor, capturing localized and semantic information~(e.g., \cite{tumanyan2022splicing,amir2021deep}). Here, we propose a \emph{new self-supervised framework that jointly and densely aligns the images in DINO-ViT features space}. To the best of our knowledge, we are the first to harness the power of DINO-ViT for \emph{dense} correspondences between in-the-wild images. More specifically, given an input image set, our  framework estimates, at test-time: (i) a joint latent 2D atlas that represents the mode of DINO-ViT features across the images, and (ii) dense mappings from the atlas to each of the images. Our training objective is driven by a matching loss encouraging each image features to match the canonical learned features in the joint atlas. We further incorporate additional loss terms that allow our framework to robustly represent and align only the shared content in the presence of background clutter or other distracting objects.  

Since our atlas and mappings are optimized per set, our method works in a zero-shot manner and can be applied to a plethora of image sets, including sets of mixed domains (e.g., aligning images depicting sculpture and artwork of cats), sets depicting related yet different object categories (e.g., dogs and tigers), or domains for which a dedicated generator is not available (e.g., coffee mugs).  We thoroughly evaluate our method, and demonstrate that our test-time optimization framework performs favorably compared to \cite{peebles2022gan} and on-par with state-of-the-art self-supervised methods. We further demonstrate how our atlas and mappings can be used for editing the image set with minimal effort by automatically propagating edits that are applied to a single image to the entire image set.

\section{Related Work}
\label{sec:relatedwork}

\paragraph{Joint Image Set Alignment.}
Congealing \cite{miller2000learning,huang2007unsupervised,learned2005data} introduced the task of jointly aligning images into a common 2D image, representing the geometric mode of the set.  This has been done by  minimizing the entropy of intensity values in a pixel stack after the alignment. They have demonstrated the use of their method for several applications, including  image classification given a few labels per class. These seminal works were extended by incorporating deep learning, combining unsupervised alignment with unsupervised feature learning \cite{huang2012learning}, which showed improvement in face verification accuracy. Further, various methods have been proposed to generalize congealing to several modes, e.g., through clustering~\cite{frey2003transformation,frey1999estimating,mattar2012unsupervised,monnier2020deep,liu2009simultaneous,vedaldi2008joint}, and make it more robust to occlusions~\cite{peng2012rasl}. 
Other methods are based on pairwise optical flow, requiring  either consistent matching across image pairs in a collection \cite{zhou2015flowweb,rubinstein2013unsupervised}, or factorizing the collection into simpler subspaces, allowing to simplify the matching task~\cite{kemelmacher2012collection,mobahi2014compositional}. AverageExplorer \cite{zhu2014averageexplorer} presented a user interactive framework for browsing Internet photo collections through average images representing modes in the collections. 

Recently, GANgealing \cite{peebles2022gan} used a Spatial Transformer Network (STN) \cite{jaderberg2015spatial} to predict a transformation from any image sampled from a predefined domain 
(e.g. cat images) into a shared aligned space. They leverage the style-pose disentanglement in a pre-trained StyleGAN2 model \cite{karras2020analyzing} to provide training supervision. Specifically, during training, their method simultaneously learns the mode of object pose across a large collection of images, and trains the STN to map each image to this mode. Their method demonstrates impressive results on complex data such as LSUN \cite{yu2015lsun}, yet requires extensive compute and large-scale training data. In contrast, we take the congealing task to the realm of test-time optimization, where only a small test image set is available (e.g., $<$25 images). Thus, our method can be applied on diverse domains, or image sets that comprise of images from mixed yet related domains, as shown in Fig.~\ref{fig:teaser}, \ref{fig:results}, \ref{fig:comp-unique-cats} and \ref{fig:comparison}.

\paragraph{Semantic Correspondences.} 
Prior to the deep-learning era, various methods have been proposed  to tackle the  task of establishing sparse point correspondences between an image pair \cite{lowe2004distinctive,tola2009daisy}. However, due to their local nature, and the lack of global context, they cannot handle significant color and shape variations.  Later, data-driven based descriptors  opened the door to establishing correspondences based on higher level information by learning from data representations that encode semantic global information; such methods either work by extracting features from a pre-trained classification model (e.g., \cite{fischer2014descriptor,simo2015discriminative,aberman2018neural}), or by training a model end-to-end for establishing semantic correspondences  (e.g., \cite{rocco2018end,rocco2018neighbourhood}). Many of them use groundtruth supervision for training \cite{cho2021cats}, while others aim to tackle the task in a weakly-supervised \cite{rocco2018end,rocco2018neighbourhood} or entirely unsupervised fashion \cite{amir2021deep}. Our method also aims  at learning semantic correspondences, however, our focus is on aligning multiple images jointly by  leveraging descriptors extracted from a pre-trained DINO-ViT model \cite{caron2021emerging}.

\paragraph{DINO-ViT Features as Local Semantic Descriptors.}

Recent works showed the power of ViT (Vision Transformer) features as local and global semantic descriptors, specifically features of a pre-trained DINO-ViT \cite{caron2021emerging}. Several works \cite{caron2021emerging,simeoni2021localizing,wang2022self} showed the use of these features for various applications such as instance segmentation, object discovery or transfer learning on downstream tasks.
GCD \cite{vaze2022generalized} categorize all unlabeled images given only a partially labeled dataset.
Other works used the pre-trained features for both object localization and segmentation \cite{melas2022deep}, and even object part discovery and segmentation \cite{choudhury2021unsupervised}.
STEGO \cite{hamilton2022unsupervised} showed a method for unsupervised semantic segmentation, which unlike most works that learn a feature per pixel, they use contrastive loss that produces low rank representation for the DINO features. 

It was shown by \cite{amir2021deep} that the deep features of a pre-trained DINO-ViT encode semantic information at fine spatial granularity and capture semantic object parts. Furthermore, they showed that this information is shared across domains of semantically-related categories, e.g. cats and dogs, allowing them to design dense descriptors which are used for various applications of co-segmentation, part co-segmentation and point correspondences between images. Splice \cite{tumanyan2022splicing} established a method for semantic appearance transfer using DINO-ViT features, and presented additional powerful properties such as keys inversion, which shows the amount of details they hold of the original image.

We build upon the findings from \cite{amir2021deep,tumanyan2022splicing} and use DINO-ViT's spatial features as dense descriptors for aligning images from semantically-related categories.

\section{Neural Atlas Congealing}
The input to our method is a collection of images that share a common semantic content, e.g., a set of images depicting different types of guitars in natural scenes.
The shared content may significantly differ across the images  in appearance, structure, pose, and appear in cluttered scenes containing complex backgrounds or other distracting objects (e.g., Fig.~\ref{fig:results}). Our goal is to automatically detect the common content across the images and estimate a geometric transformation that maps each image into a joint 2D space. Our key idea is to harness the power of deep features extracted from a pre-trained (and fixed) DINO-ViT model, which have been shown to capture localized semantic information under significant appearance and pose variations~\cite{caron2021emerging,amir2021deep,tumanyan2022splicing}.  

\begin{figure*}[t!]
    \centering
    \includegraphics[width=\textwidth]{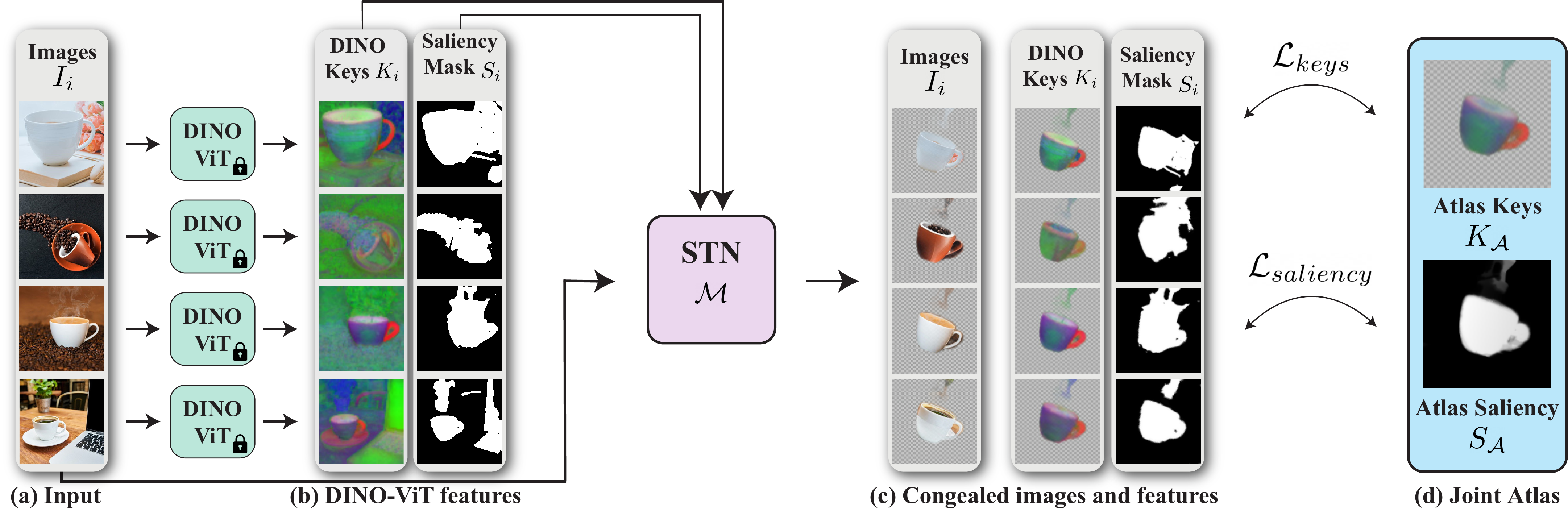}
    \caption{\emph{Neural Congealing.} Each of the input images $I_i$ in the set (a) is fed into a pre-trained and fixed DINO-ViT model. (b) Spatial features, $K_i$, are extracted and processed to estimate initial rough saliency masks $S_i$ (see \cite{amir2021deep}), both of which are used to train: (i) a mapping network (STN), which (c) transforms each of the input images into a joint 2D space, and (ii) (d) a joint learnable 2D feature atlas, $K_{\mathcal{A}}$, and an atlas saliency map $S_\mathcal{A}$. 
    Note that while the input saliency masks are noisy and contain objects from the background, our method is able to refine them and focus on the most common object in the set. See SM for the full set of images.}
    \label{fig:pipeline}\afterfigure
\end{figure*}

Specifically, our framework, illustrated in Fig.~\ref{fig:pipeline}, jointly aligns the images in DINO-ViT feature space using two  learnable components: (i) a unified latent 2D atlas that represents the common semantic mode in DINO-ViT space across the images, and (ii) a Spatial Transformer Network (STN) that aligns each of the input images to the joint latent atlas. Our method is fully automatic and self-supervised: each of the input images is first fed into DINO-ViT and features (keys) are extracted from the last layer, and serve as our spatial semantic descriptors. The atlas representations, in which each  pixel stores a latent feature, and the STN parameters are then optimized such that the transformed features of each image are aligned with the joint atlas features. 

We further define a saliency value in each atlas pixel that takes a continuous value between zero and one. We optimize the atlas saliency using a voting-based loss w.r.t. rough initial image saliency masks that are estimated from DINO-ViT features  in a pre-processing step~\cite{amir2021deep}.  This approach allows us to robustly align only the common regions in highly cluttered scenes.

\subsection{Semantic Joint Image Alignment}\label{sec:semantic-alignment}

Given an input image set $\{I_i\}_{i=1}^N$ and a pre-trained (and fixed) DINO-ViT model, we extract for each image its keys features from the last layer, denoted by $K_i \in \mathbb{R}^{H\times W \times D}$;
an initial per-image saliency mask $S_i$  is estimated by applying  a simple clustering-based method directly to the extracted features (see~\cite{amir2021deep}). These  maps roughly capture salient foreground regions, but are often noisy and contain uncommon regions across the images, as seen in Fig.~\ref{fig:refined-masks}.

We define a learnable atlas $\mathcal{A}$ as a 2D grid of latent features $K_\mathcal{A}\in \mathbb{R}^{H_\mathcal{A} \times W_\mathcal{A} \times D}$, and a saliency mask $S_\mathcal{A}\in \mathbb{R}^{H_\mathcal{A} \times W_\mathcal{A}} $.  
We define the 2D mapping of each atlas point $\vv{x}_{\mathcal{A}}$ to each of the input images $I_i$ as follows:
    \begin{equation}
    \vv{x}_i=\mathcal{M}(I_i,\vv{x}_\mathcal{A})
    \label{eq:mapping}
\end{equation}
where $\vv{x}_i$ is the estimated corresponding point of $\vv{x}_\mathcal{A}$ in image $I_i$.
Note that applying $\mathcal{M}$ on each of the atlas coordinates allows us to backward warp $I_i,S_i$ or $K_i$ into the atlas space.

Similarly to \cite{peebles2022gan}, $\mathcal{M}$ is modeled as a composition of rigid and non-rigid transformations $\mathcal{M}_r \circ \mathcal{M}_f$, each is estimated by a separate STN. 
That is,  $\mathcal{M}_r$ is a global 2D similarity transformation defined by:  

\begin{equation}
    \mathcal{M}_r(I_i,\vv{x}) = sR\vv{x}+\vv{t}
\end{equation}  
where  $R\in SO(2)$,  $\vv{t} \in \mathbb{R}^2$ and $s\in \mathbb{R}^+$ are 2D rotation, translation and global scale, respectively. Our method also supports horizontal flips (see Appendix~\ref{app:ext-d}).

Given an image $I$, the non-rigid transformation $\mathcal{M}_f$ is  defined by a dense flow field: 
\begin{equation} \label{eq::flow_transformation}
     \mathcal{M}_f(I,\vv{x}) = \vv{x}+\vv{w}
\end{equation}
where $\vv{w}$ is a per-pixel 2D offset. 

In practice, each of the transformations is obtained by an STN model: one takes $I_i$ as input and predicts the parameters of the similarity transformation $\mathcal{M}_r$; the  image  $I_i$ is then backward warped using $\mathcal{M}_r$ and fed to the second STN that predicts $\mathcal{M}_f$. See network architecture details in Appendix~\ref{app:stn-arch}.

\subsection{Training}\label{sec:training} Given $\{I_i, S_i, K_i\}_{i=1}^N$,  we now turn to the task of learning the joint atlas  $\mathcal{A}=(K_\mathcal{A}, S_\mathcal{A})$ and 2D mappings $\mathcal{M}=(\mathcal{M}_r, \mathcal{M}_f)$. Our objective function incorporates four main loss terms and takes the following form: 
\begin{equation}
\mathcal{L} = \mathcal{L}_{keys} + \lambda_s\mathcal{L}_{saliency} +  \lambda_r\mathcal{L}_{reg_{\mathcal{M}}} + \lambda_a\mathcal{L}_{reg_\mathcal{A}} 
\label{eq:objective}
\end{equation}
where $\lambda_s, \lambda_a$ and $\lambda_r$ control the relative weights between the terms, and are fixed throughout training. 

\paragraph{Semantic loss ${\bf \mathcal{L}_{keys}}$.} 
The semantic loss is our driving loss and it encourages an alignment in DINO-ViT feature space between the atlas and each of the congealed images: 
{\small \begin{equation}
\begin{aligned}
    & \mathcal{L}_{keys} = \sum_{i=1}^N  \frac{1}{N \cdot \Sigma_{S_{\mathcal{A}}}} \sum_{\vv{x}_\mathcal{A}} S_\mathcal{A}(\vv{x}_\mathcal{A}) \cdot   \mathcal{D}\left(K_i(\vv{x}_i),K_{\mathcal{A}}(\vv{x}_\mathcal{A})\right) \\
\end{aligned}\label{eq:keys-loss}
\end{equation}}
where  $\Sigma_{S_{\mathcal{A}}} = \sum_{\vv{x}_\mathcal{A}} S_\mathcal{A}(\vv{x}_\mathcal{A})$, $K_\mathcal{A}(\vv{x}_\mathcal{A})$ is the atlas feature at location $\vv{x}_\mathcal{A}$, and $K_i(\vv{x}_i)$ is the corresponding feature in $I_i$ (see Eq.~\ref{eq:mapping}). The distance metric $\mathcal{D}$ is defined by: $ \lambda_l\norm{K_i(\vv{x}_i)-K_{\mathcal{A}}(\vv{x}_\mathcal{A})}_2^2+\mathcal{D}_{cos}\left(K_i(\vv{x}_i), K_{\mathcal{A}}(\vv{x}_\mathcal{A})\right)$, where $\mathcal{D}_{cos}$ is the cosine distance. Note that since we are interested in aligning only the semantically common content across the images, our loss is weighted according to the atlas saliency $S_\mathcal{A}$.

\paragraph{Saliency loss ${\bf \mathcal{L}_{saliency}}$.}  This term serves as a saliency-voting loss that allows us to capture the common content across the images by the atlas saliency $S_\mathcal{A}$. Formally, the atlas saliency is optimized to match the initial congealed images' saliency masks. Since the image saliency masks are often rough and contain clutter or irrelevant salient objects of the scene (Fig.~\ref{fig:pipeline}), we use a robust loss:

\begin{equation}
\begin{aligned}
    & \mathcal{L}_{saliency} = \frac{1}{N \cdot N_\mathcal{A}}  \sum_{i=1}^N  \sum_{\vv{x}_\mathcal{A}}  \rho_{0.7}\left(S_i(\vv{x}_i), S_\mathcal{A}(\vv{x}_\mathcal{A})\right) 
\end{aligned}
\end{equation}
where $N_\mathcal{A}$ is the number of pixels in the atlas, and $\rho_\delta(a,b)$ denotes the Huber loss \cite{huber} with parameter $\delta$:
\begin{equation}
  \rho_\delta(a,b)= \begin{cases} \begin{array}{lr}
        \frac{1}{2} (a-b)^2, & \text{if } |a-b|<\delta \\
         \delta \cdot \left( |a-b| - \frac{1}{2} \delta \right)  & \text{otherwise }
        \end{array} 
        \end{cases}
\end{equation}
Intuitively, each image ``votes'' for the regions that should be salient in the atlas, and the aggregated common salient content is estimated. 

\paragraph{Transformation regularization ${\bf \mathcal{L}_{reg_\mathcal{M}}}$.} 
For obtaining a shared representation that is as undistorted as possible, while containing some distortions for aligning objects with different proportions of semantic parts, we apply regularization on both mapping networks:
\begin{equation}
\mathcal{L}_{reg_\mathcal{M}} = \lambda_{s_1}\mathcal{L}_{scale} + \lambda_{s_2}\mathcal{L}_{mag} +\mathcal{L}_{smooth}
\end{equation}
where $\lambda_{s_1}$ and $\lambda_{s_2}$ are the relative weights.

$\mathcal{L}_{scale}$ regularizes $\mathcal{M}_r$ from changing the scale of the original images in the atlas space: 
\begin{equation}
\mathcal{L}_{scale} = \frac{1}{N} \sum_{i=1}^N \vert 1-s_i\vert^2
\end{equation} 
where $s_i$ is the scale parameter of the learned rigid transformation for image $I_i$.

The non-rigid transformation is encouraged to be as small as possible:
\begin{equation}
\mathcal{L}_{mag} = \frac{1}{N\cdot N_\mathcal{A}}\sum_{i=1}^N \sum_{x_\mathcal{A}}  \norm {\vv{w}_i}_2^2 
\end{equation}
where $\vv{w}$ is the per-pixel flow vector defined in Eq.~\eqref{eq::flow_transformation}.
The term $\mathcal{L}_{smooth}$, defined as in \cite{kasten2021layered}, is used to prevent the non-rigid mapping from distorting the shared content by encouraging as rigid as possible mapping.  Formally, this term is defined by:
 \begin{equation}
\mathcal{L}_{smooth} = \frac{1}{N\cdot N_\mathcal{A}}\sum_{i=1}^N \sum_{x_\mathcal{A}}  \left(\norm{J^TJ}_F+\norm{\left(J^TJ\right)^{-1}}_F\right)
\end{equation}
where  $J$ is the Jacobian matrix of $\mathcal{M}$ at $\vv{x}_\mathcal{A}$. See Appendix~\ref{app:loss-terms} for more details.

\begin{figure*}[t!]
    \centering
    \includegraphics[width=1\textwidth]{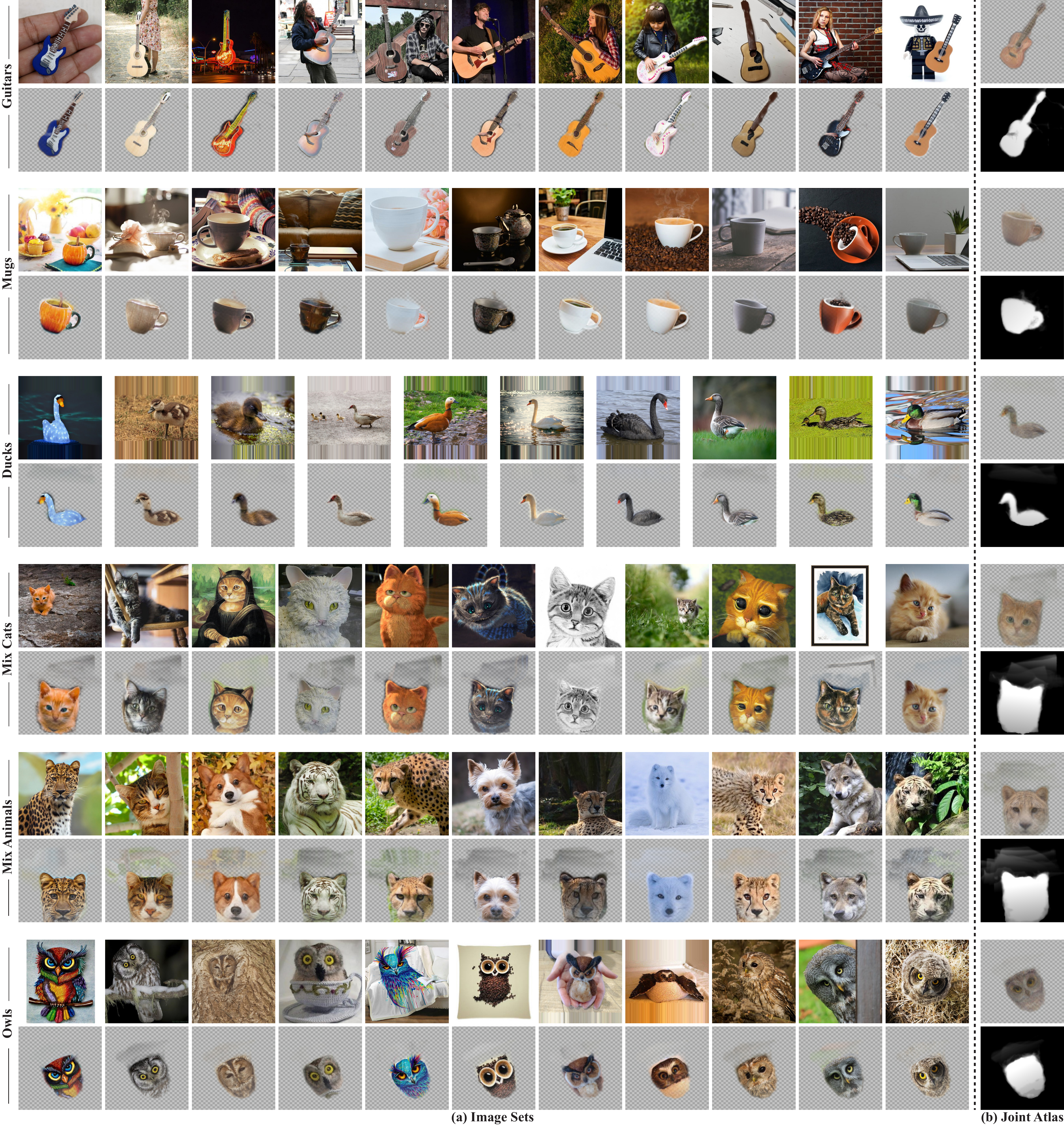}
    \caption{ \emph{Neural Congealing results.} We show results on a variety of image sets, each containing images with large variations in pose, appearance, scale, different domains, and cluttered backgrounds. We show (a) the original images together with the congealed images, and (b) the average image in atlas space together with the joint atlas saliency mask. See SM for results on all full sets. }
    \label{fig:results}\afterfigure
\end{figure*}

\paragraph{Atlas regularization ${\bf \mathcal{L}_{reg_\mathcal{A}}}$.}  This term controls the localization and sparsity of the atlas:
\begin{equation}
\mathcal{L}_{reg_\mathcal{A}} = \mathcal{L}_{center} + \lambda_p\mathcal{L}_{sparsity}
\end{equation}
where $\lambda_p$ is the relative weight.

Since the position of the shared content in the atlas space is arbitrary, we define $\mathcal{L}_{center}$ to encourage the common object to be mapped to the center of the atlas space: 
\begin{equation}
\mathcal{L}_{center} = \norm{\frac{1}{\Sigma_{S_{\mathcal{A}}}} \cdot \sum_{\vv{x}_\mathcal{A}} S_\mathcal{A}(\vv{x}_\mathcal{A}) \cdot \vv{x}_\mathcal{A}}_2^2
\end{equation}
where $\vv{x}_\mathcal{A}$ coordinates are normalized to be in the range $(-1,1)\times (-1,1)$. By minimizing the norm of the saliency's center of mass, we encourage it to be as close as possible to the atlas center that is located at the origin.

We further observe that without any sparsity regularization, the atlas often contains non-common information. 
$\mathcal{L}_{sparsity}$ encourages both $S_\mathcal{A}$ and $K_\mathcal{A}$ to be sparse:
\begin{equation}
\mathcal{L}_{sparsity} =  \mathcal{L}_{sparsity}^{S_\mathcal{A}} + \lambda_{k}\mathcal{L}_{sparsity}^{K_\mathcal{A}}
\end{equation}
where $\lambda_{k}$ is the relative weight. We follow \cite{retiming,omnimatte,bar2022text2live} and define the sparsity loss term for the atlas saliency as a combination of L1- and L0-approximation regularization terms 
\begin{equation}
\mathcal{L}_{sparsity}^{S_\mathcal{A}} = \gamma \norm {S_\mathcal{A}}_1 + \Psi_0(S_\mathcal{A})
\end{equation}
where $\Psi_0(x) \equiv 2Sigmoid(5x) - 1$ is a smooth L0 approximation that penalizes non zero elements, and $\gamma$ is the relative weight between the terms.

For the atlas features, we apply L1 sparsity loss on non-salient parts only:
\begin{equation}
\mathcal{L}_{sparsity}^{K_\mathcal{A}} = (1 - S_\mathcal{A})\cdot\norm {K_\mathcal{A}}_1
\end{equation}

\subsection{Editing}\label{sec:editing}
Once we have the atlas representation, we can use the average image of all congealed images as a template for editing. Then, the edit in the atlas space is automatically propagated back to all original images. One can also apply an edit on one of the images and propagate it to all the rest by passing through the atlas space. 

As in \cite{peebles2022gan}, given an image $I_i$ and an RGBA edit image in atlas space we apply forward warping using $\mathcal{M}(I_i,\vv{x}_\mathcal{A})$ to the image space and then apply alpha blending of the warped edit image with $I_i$. 

\section{Results}
We tested our method on a variety of image sets, containing 5-25 images, from LSUN dataset \cite{yu2015lsun}, AFHQ dataset \cite{choi2020stargan}, Pixabay \cite{pixabay}, Shutterstock \cite{shutterstock},
and the general Internet. Each set contains objects that share the same semantic parts from different categories including painted/animated/real animals, mugs, guitars, etc. The domain, pose/orientation and appearance, as well as the amount of irrelevant salient objects in the background, may significantly change between the images in each set. We pre-process the images to $256\times 256$ px, with border padding in case of an unsuited aspect-ratio. See Appendix~\ref{app:impl-details} for full implementation details.

\subsection{Qualitative Results.}
\begin{figure}[t!]
    \centering
    \includegraphics[width=\columnwidth]{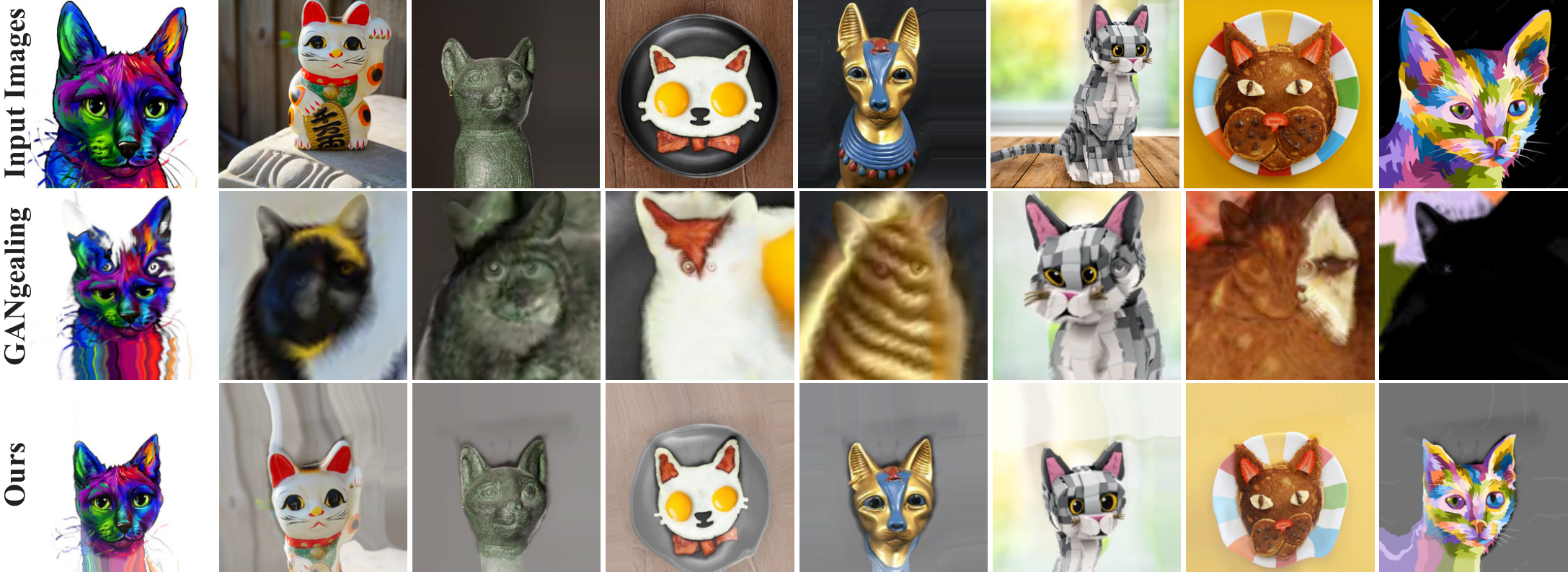}
    \caption{\emph{Across-domain set.} We compare results to GANgealing \cite{peebles2022gan}. While they struggle with such images, our method is able to semantically align them.}
    \label{fig:comp-unique-cats}\afterfigure
\end{figure}

\begin{figure*}[t!]
    \centering
    \includegraphics[width=1\textwidth]{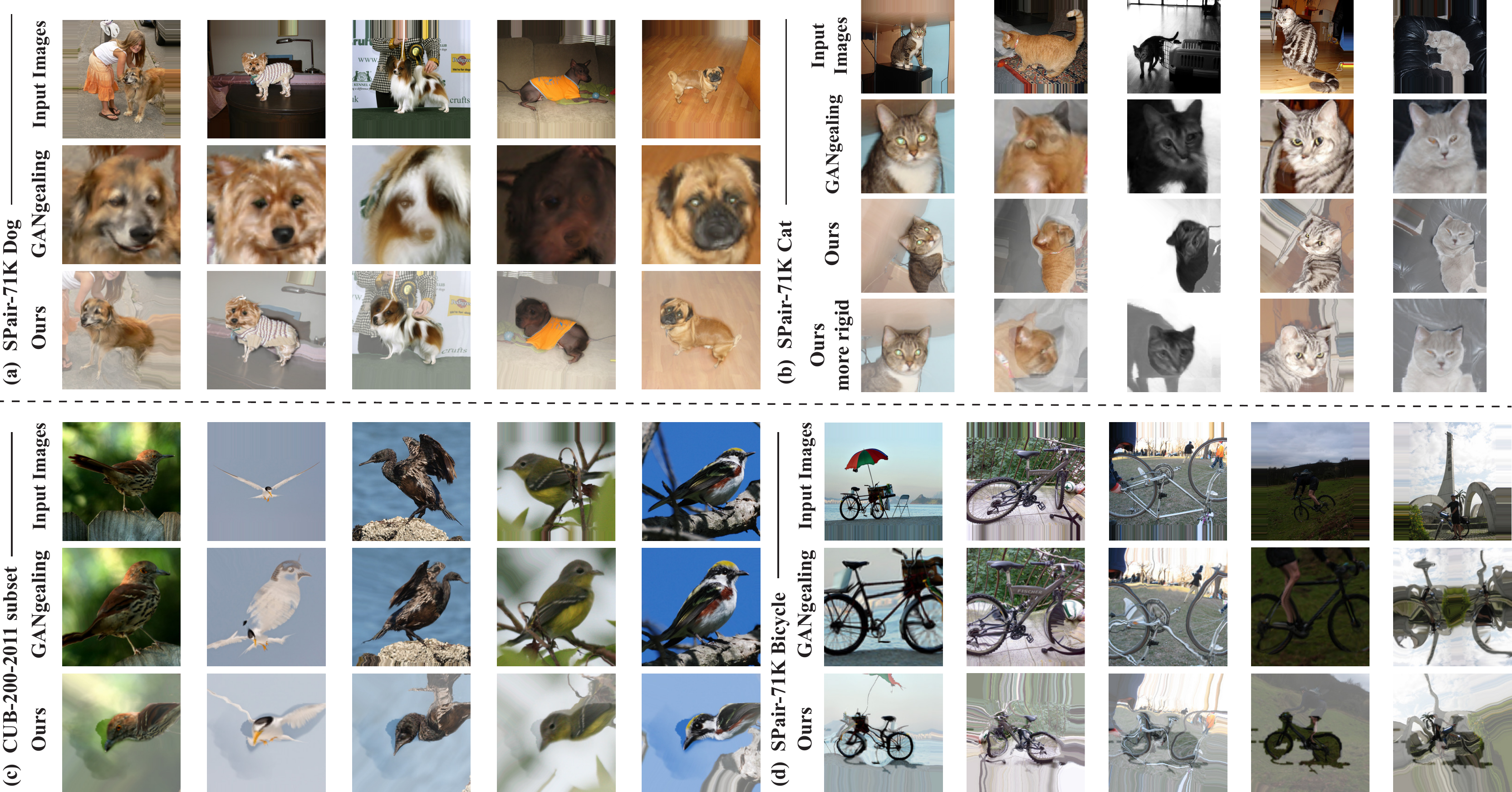}
    \caption{\emph{Comparison to GANgealing \cite{peebles2022gan}.} We compare results from sets used in evaluation (Sec.~\ref{sec:quantitative}). Since our method is based on alignment according to semantic DINO-ViT features, even in hard examples, our method focuses on aligning the most common salient object part, e.g., the birds in (c). Note that as in \cite{peebles2022gan}, our method supports horizontal flips (see Appendix~\ref{app:ext-d} for technical details).
    }
    \label{fig:comparison}\afterfigure
\end{figure*}

Sample examples for input sets along with our joint alignment result can be seen in Fig.~\ref{fig:teaser}, \ref{fig:pipeline}, \ref{fig:results}, \ref{fig:comp-unique-cats} and \ref{fig:comparison}. Fig.~\ref{fig:results} also shows visualizations of the average image in the atlas space and the atlas saliency. The full set of results is included in the Supplementary Materials (SM). As can be seen, our method successfully aligns diverse in-the-wild sets under significant  differences in object scale  (e.g. \emph{Mugs} in Fig.~\ref{fig:results}), proportions between semantic parts (e.g., Garfield's ears in \emph{Mix Cats}, or the Corgi's ears in \emph{Mix Animals} in Fig.~\ref{fig:results}), slight differences in out-of-plane rotation (e.g., white tiger in \emph{Mix Animals} in Fig.~\ref{fig:results}), and under non-rigid deformations (e.g., the butterflies in Fig.~\ref{fig:teaser}). 

In addition, Fig.~\ref{fig:comp-unique-cats} shows an example where our method aligns images across different domains (paintings, food, etc.). This demonstrates the flexibility of our test-time training approach compared to GANgealing \cite{peebles2022gan} that was trained on naturally-looking images, and thus struggles to generalize to other domains.

\begin{figure}[t!]
    \centering
    \includegraphics[width=\columnwidth]{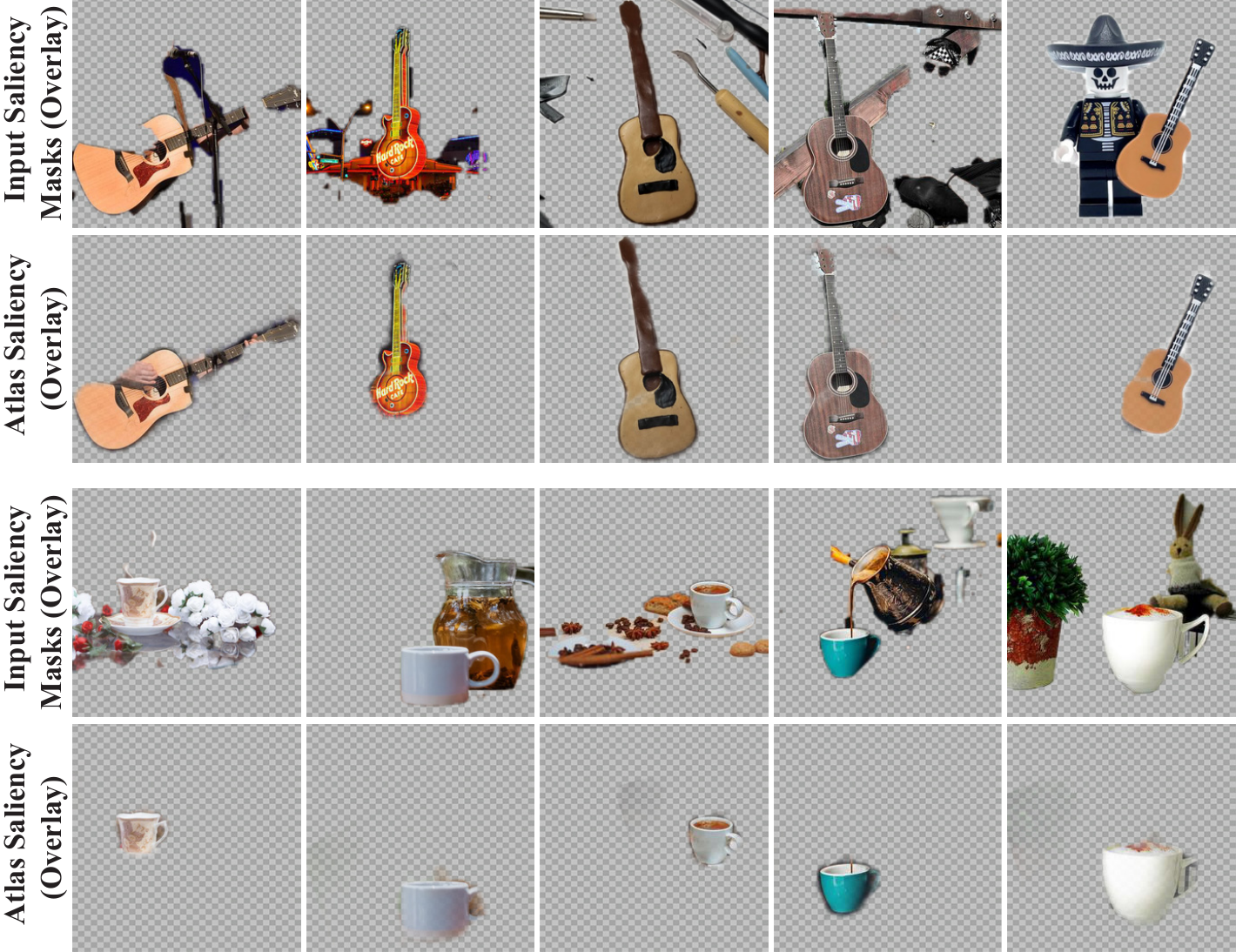}
    \caption{\emph{Refined saliency masks (overlay).} Initial saliency masks extracted using~\cite{amir2021deep}, together with our final learned atlas saliency mask. Our method is able to refine the masks capturing the common object, while ignoring cluttered background content or irrelevant objects.}
    \label{fig:refined-masks}\afterfigure
\end{figure}

\begin{figure}[t!]
    \centering
    \includegraphics[width=\columnwidth]{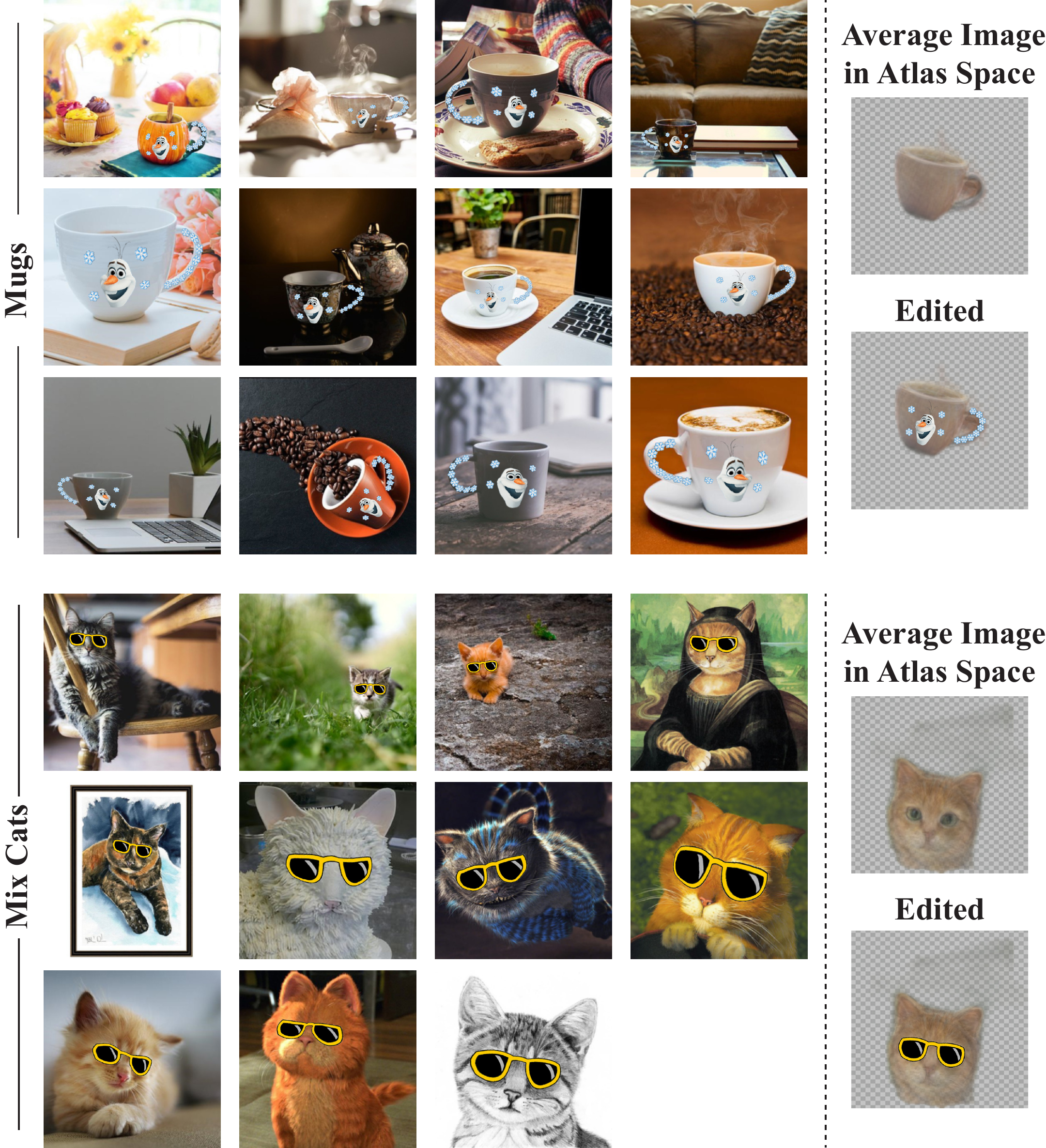}
    \caption{\emph{Editing results.} Sample results for editing for two different sets. Edits are propagated to the same semantic parts to all images. See SM for more results.}
    \label{fig:edits}\afterfigure
\end{figure}

\paragraph{Edit results.} Fig.~\ref{fig:teaser} and \ref{fig:edits} show sample editing results automatically applied to the input set (Sec.~\ref{sec:editing}). As can be seen, the edits are mapped correctly and accurately to the same semantic regions in all images, under significant variation in scale, pose and appearance. More editing results are included in the SM.

\paragraph{Refined masks.} As discussed in Sec.~\ref{sec:semantic-alignment}, the initial saliency masks extracted from  DINO-ViT features using \cite{amir2021deep} are typically very coarse and may contain irrelevant content such as other objects. Fig.~\ref{fig:pipeline} and \ref{fig:refined-masks} show examples of how our method manages to congeal these rough estimates into an accurate and refined mask that captures the shared content, while robustly filtering out cluttered background content or non-shared objects. The full set of saliency masks is included in the SM.

\subsection{Quantitative Evaluation and Comparison.} \label{sec:quantitative}
We evaluate our framework on the task of semantic point correspondences on SPair-71K \cite{min2019spair} and CUB-200-2011 \cite{WahCUB_200_2011}. Specifically, given a source image $I_A$ and a target image $I_B$ together with their ground truth point correspondences, we transfer the points from image $I_A$ to the atlas space and map them back to image $I_B$ to obtain its predicted points (see Appendix~\ref{app:correspondences} for technical details). We then measure for each set the PCK-Transfer, i.e., the percentage of keypoints that are mapped within the threshold of $\alpha\cdot \max(h,w)$ from the ground truth.  We follow previous works and set $\alpha=0.1$ for both benchmarks, and $h, w$ to be the dimensions of the object's bounding box for SPair-71K. For CUB, we follow \cite{peebles2022gan} and set $h, w$ to be the image size. 

\paragraph{Results on SPair-71K.} We use the same pre-processing as in \cite{peebles2022gan}, applying border padding for non-square images and resizing to $256\times 256$. We apply our method on each test set separately, each includes 25-26 images.  See Appendix~\ref{app:ext-d} for further technical details.

Table \ref{tab:spair} reports the results for our method, GANgealing, and a number of leading methods for semantic correspondences. As seen,  our method outperforms GANgealing on most sets, and outperforms other self-supervised methods on all sets. 
Our method performs very well on the \emph{Cat} and \emph{Dog} sets, yet in the \emph{Bicycle} set, due to the large deformations and the symmetric shape of the object, our performance decreases. Nevertheless, even in this challenging case our performance is on-par with most supervised methods on this set. Note that all supervised methods have been directly trained or fine-tuned using ground-truth supervision on the SPair-71K training set. 

Figure~\ref{fig:comparison} shows qualitative comparisons to \cite{peebles2022gan}; GANgealing converges to a global mode across a large dataset, which allows them to achieve good alignment for \emph{Bicycle} and \emph{Birds} (Fig.~\ref{fig:comparison}(c),(d)). However, in the \emph{Dog} set this restricts their alignment to capturing only the head, while our method detects the common mode in a given set and can align the full body of the dogs (Fig.~\ref{fig:comparison}(a)).

There is an inherent tradeoff between aligning highly articulated content (e.g., the bodies of the non-rigid animals) and maintaining undistorted atlas representation. We demonstrate this tradeoff for the \emph{Cat} set, by controlling the effective relative weight of our rigidity loss. As seen, with our default parameters, the method aligns the bodies of the cats yet their faces are not accurately aligned. By increasing the relative weight of the rigidity, we can encourage the model to focus on the most rigid part across the set, while disregarding the cats' body (even though it is a shared salient part across the set). 
In this setting, fine facial details are  accurately aligned and our method outperforms all previous methods, including supervised methods.

\begin{table}
\footnotesize   
  \centering
  \begin{tabular}{@{}lcccc@{}}
    \toprule
    Method & Supervision & Cat & Dog & Bicycle \\
    \midrule
    HPF \cite{min2019hyperpixel} & supervised & 52.9 & 32.8 & 18.9 \\
    DHPF \cite{min2020learning} & supervised & 61.6 & 46.1 & 23.8\\
    SCOT \cite{liu2020semantic} & supervised & 63.1 & 42.5 & 20.7 \\
    CHM \cite{min2021convolutional} & supervised & 64.9 & 56.1 & 29.3 \\
    CATs \cite{cho2021cats}  & supervised & 66.5 & 56.5 & 34.7 \\
    \midrule
    WeakAlign \cite{rocco2018end} & weakly-supervised & 31.8 & 22.6 & 17.6 \\
    NC-Net \cite{rocco2018neighbourhood} & weakly-supervised & 39.2 & 18.8 & 12.2 \\
    \midrule
    CNNgeo \cite{rocco2017convolutional} & self-supervised & 32.7 & 22.8 & 16.7 \\
    A2Net \cite{seo2018attentive} & self-supervised &  35.6 & 24.3 & 18.5 \\
    GANgealing \cite{peebles2022gan} & GAN-supervised & 67.0 & 23.1 & \textbf{37.5} \\
    Ours & self-supervised & 54.5/\textbf{70.7}* & \textbf{35.8} & 29.1 \\
    \bottomrule
  \end{tabular}
  \caption{\emph{PCK-Transfer on categories from SPair-71K with $\alpha=0.1$.} *Results for relaxed setting, aligning only the cats' heads. }
  \label{tab:spair}
\end{table}

\paragraph{Results on CUB-200-2011.} Since our method works with small image sets, we randomly sample 14 sets of 25 images each, and train separately on each of them. For fair comparison, we apply the same pre-processing as in \cite{peebles2022gan}. 
As seen in Table \ref{tab:cub}, our method achieves better results compared to GANgealing. As seen in Fig.~\ref{fig:comparison}(c), GANgealing tends to hallucinate object parts, and struggles with aligning the heads, especially when the object pose differs significantly from the canonical pose learned from the entire domain. Our method, by optimizing the representation and mappings per set, manages to align the heads of the birds even under unusual poses, e.g., distant bird  spreading wings (second column from the left).

\subsection{Ablation Study}\label{sec:ablation} 
\begin{figure}[t!]
    \centering
    \includegraphics[width=\columnwidth]{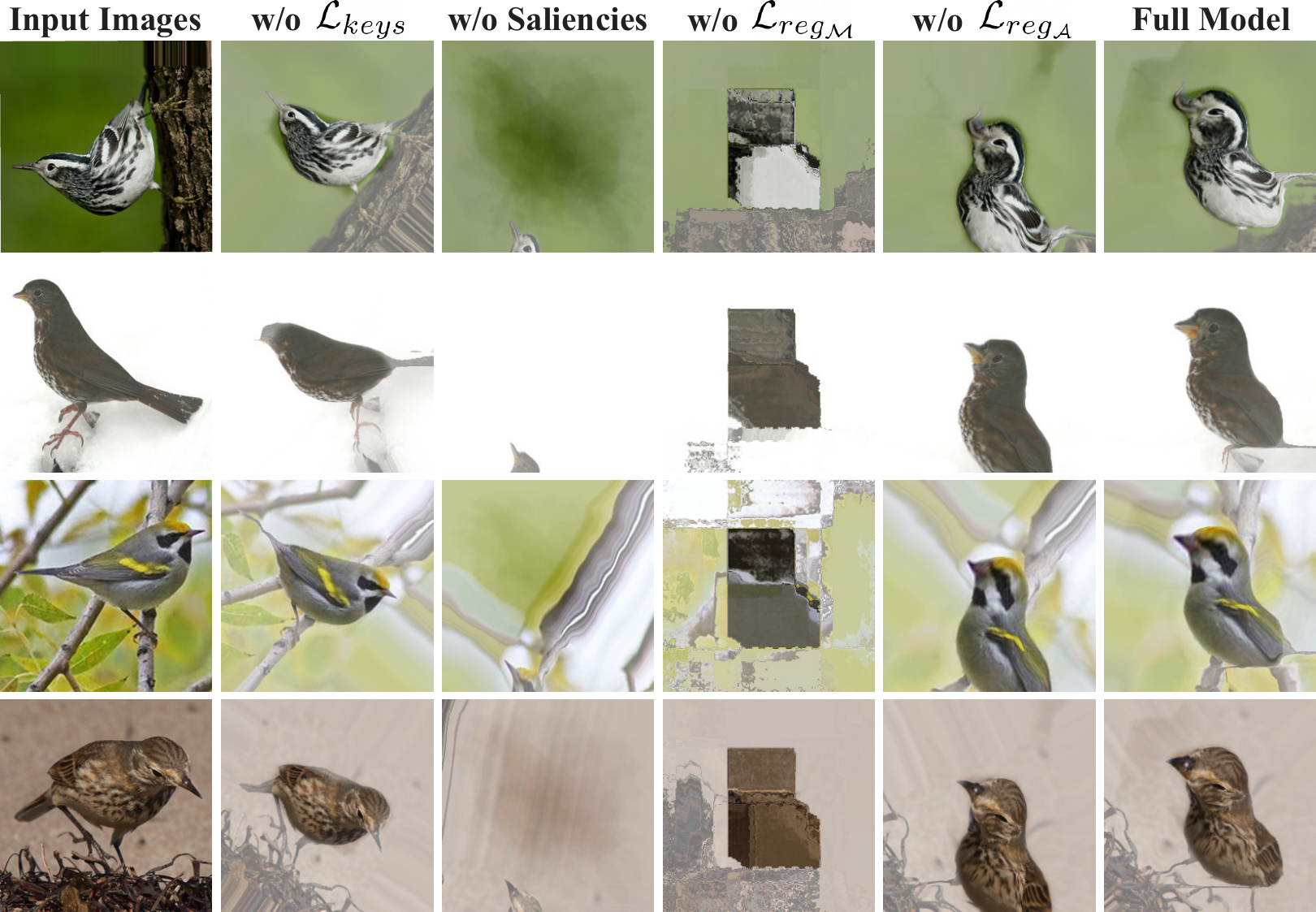}
    \caption{Ablation of each of the loss terms in our objective function. See Sec.~\ref{sec:ablation}.}
    \label{fig:ablation}\afterfigure
\end{figure}

We ablate the different loss terms of our objective function (Eq.~\ref{eq:objective}), both quantitatively in Table~\ref{tab:cub} and qualitatively in Fig.~\ref{fig:ablation}. Without our driving loss $\mathcal{L}_{keys}$,  we notice a significant drop in performance. Fig.~\ref{fig:ablation} shows that even though the saliency masks help a great deal in bringing the birds one on top of the other, there is no semantic alignment between them. Without saliency masks (no atlas saliency), our framework attempts to align all observed content, and thus struggles to converge, or converges only to small parts of the object, in cases of significant background clutter. 

Removing $\mathcal{L}_{reg_\mathcal{M}}$ provides too much freedom to the non-rigid mapping, which converges either to a single point in the atlas, or spreads in disorder. Thus, the performance drops dramatically and the visual results are not appealing. Our method performs on-par without $\mathcal{L}_{reg_\mathcal{A}}$, however, we note that the initial saliencies for the CUB subsets are quite accurate, thus there is no much need for regularizing the atlas in this case.
More generally, this loss allows us to obtain cleaner atlases (see Appendix~\ref{app:no-reg-a}). 
Finally, we see in Fig.~\ref{fig:ablation} that $\mathcal{L}_{center}$ of $\mathcal{L}_{reg_\mathcal{A}}$ encourages the shared content to be centered, allowing us to keep the birds within atlas borders.
\begin{table}
\small 
  \centering
  \begin{tabular}{@{}lcclc@{}}
    \toprule
    Method & Supervision & PCK@$\alpha=0.1$ \\
    \midrule
    Ours w/o $\mathcal{L}_{keys}$ & self-supervised & 30.7 \\
    Ours w/o saliency masks & self-supervised & 53.4 \\
    Ours w/o $\mathcal{L}_{reg_\mathcal{M}}$ & self-supervised & 36.9 \\
    Ours w/o $\mathcal{L}_{reg_\mathcal{A}}$ & self-supervised & \textbf{64.9} \\
    \midrule
    GANgealing \cite{peebles2022gan} & GAN-supervised & 56.8 \\
    Ours & self-supervised & \textbf{63.6} \\
    \bottomrule
  \end{tabular}
  \caption{\emph{PCK-Transfer on subsets of CUB-200-2011.} Comparison to \cite{peebles2022gan}, together with ablation for our different loss terms. }
  \label{tab:cub}
\end{table}

\section{Limitations}
\begin{figure}[t!]
    \centering
    \includegraphics[width=\columnwidth]{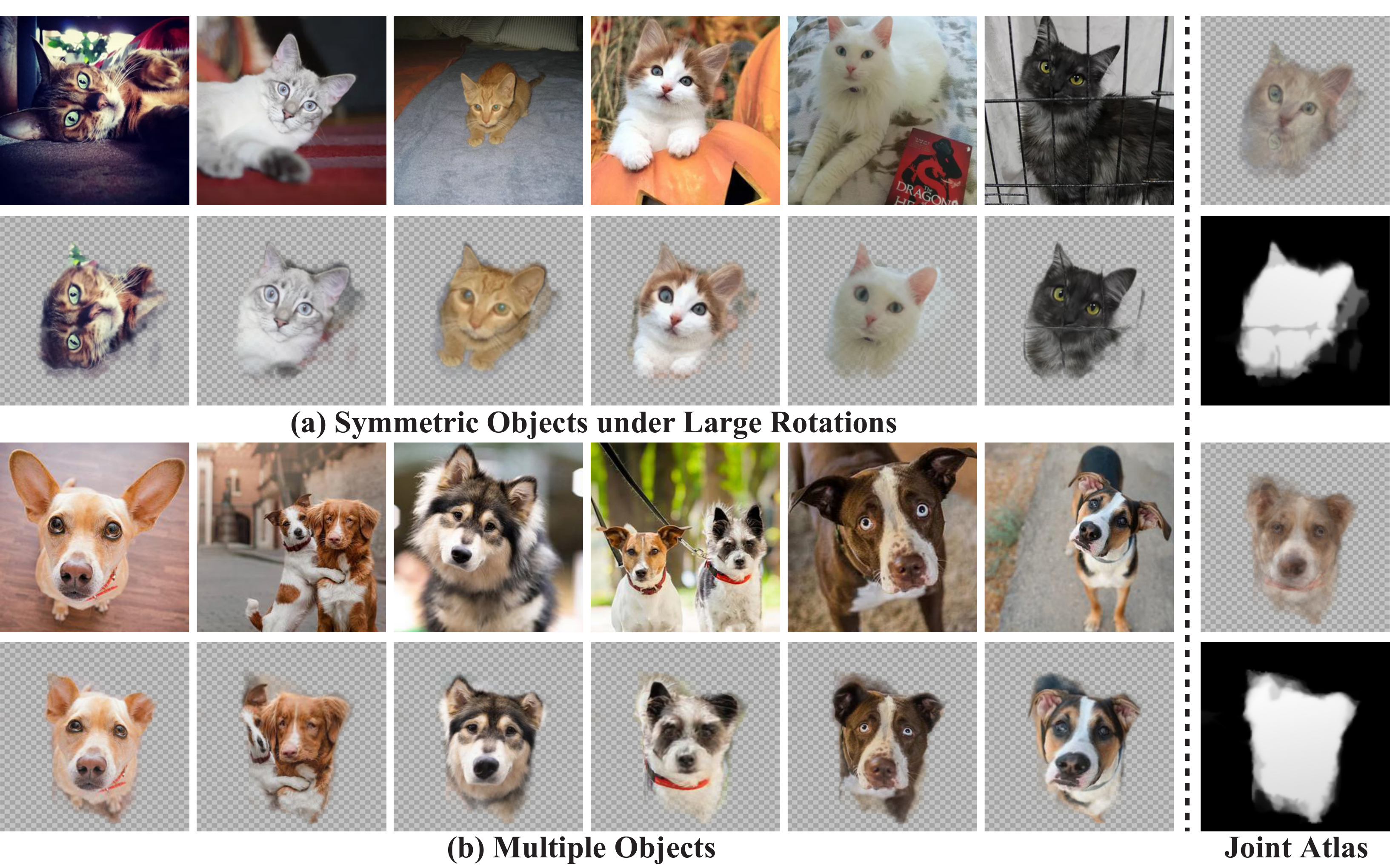}
    \caption{\emph{Limitations.} Sets containing symmetric objects under large rotations may converge partially due to relative position between semantic parts. In addition, our method is not designed to align images depicting more than one instance of the shared mode.}
    \label{fig:limitations}\afterfigure
\end{figure}

Our method relies on semantic similarity in the space of DINO-ViT feature space. Hence, in cases where these features do not capture the semantic association across the images, our method would not work well (e.g., image domains that are not well-represented in DINO's training data). 
In addition, in cases of extreme topological changes in the common object across the set, our method struggles to converge to a good alignment due to the strong rigidity constraints, e.g., Fig.~\ref{fig:comparison}(b) and Fig.~\ref{fig:comparison}(c).
Furthermore, we notice that in sets containing symmetric objects with large rotation differences, the relative position between parts may affect the convergence, and may lead to partial alignment, e.g., Fig.~\ref{fig:limitations}(a), where the left eye of the leftmost cat is matched to the right eye of the rest of the cats.
In general, our framework is not designed to align images depicting more than one instance of the shared mode. In this case, our method may align arbitrarily one of the objects, and in other cases may fail to converge, e.g., Fig.~\ref{fig:limitations}(b).

\section{Conclusions}
We tackled the congealing task in a particularity challenging setting --  jointly aligning a small set of in-the-wild images,  without any additional training data other than the test set itself. We showed how to leverage the power of pre-trained DINO-ViT features for this task in a new test-time training framework. We demonstrated the key advantages of our approach w.r.t. existing state-of-the-art methods in its applicability to diverse image domains, lightweight training and overall performance.  We further showed that our method can be used for automatically propagating edits to the entire set by simply editing a single image.  We believe that our approach -- combining test-time optimization with semantic information learned by external large-scale models -- holds great promise for dense alignment tasks, and can motivate future research in this direction.  

\paragraph{Acknowledgements:} We thank Shai Bagon and Shir Amir for their insightful comments. This project received funding from the Israeli Science Foundation (grant 2303/20).

{\small
\bibliographystyle{ieee_fullname}
\bibliography{egbib}

\begin{thebibliography}{10}\itemsep=-1pt

\bibitem{aberman2018neural}
Kfir Aberman, Jing Liao, Mingyi Shi, Dani Lischinski, Baoquan Chen, and Daniel
  Cohen-Or.
\newblock Neural best-buddies: Sparse cross-domain correspondence.
\newblock {\em ACM Transactions on Graphics (TOG)}, 2018.

\bibitem{amir2021deep}
Shir Amir, Yossi Gandelsman, Shai Bagon, and Tali Dekel.
\newblock Deep vit features as dense visual descriptors.
\newblock {\em ECCVW What is Motion For?}, 2022.

\bibitem{bar2022text2live}
Omer Bar-Tal, Dolev Ofri-Amar, Rafail Fridman, Yoni Kasten, and Tali Dekel.
\newblock Text2live: Text-driven layered image and video editing.
\newblock In {\em Computer Vision--ECCV 2022: 17th European Conference, Tel
  Aviv, Israel, October 23--27, 2022, Proceedings, Part XV}, pages 707--723.
  Springer, 2022.

\bibitem{caron2021emerging}
Mathilde Caron, Hugo Touvron, Ishan Misra, Herv{\'e} J{\'e}gou, Julien Mairal,
  Piotr Bojanowski, and Armand Joulin.
\newblock Emerging properties in self-supervised vision transformers.
\newblock In {\em Proceedings of the IEEE/CVF International Conference on
  Computer Vision}, pages 9650--9660, 2021.

\bibitem{cho2021cats}
Seokju Cho, Sunghwan Hong, Sangryul Jeon, Yunsung Lee, Kwanghoon Sohn, and
  Seungryong Kim.
\newblock Cats: Cost aggregation transformers for visual correspondence.
\newblock {\em Advances in Neural Information Processing Systems}, 2021.

\bibitem{choi2020stargan}
Yunjey Choi, Youngjung Uh, Jaejun Yoo, and Jung-Woo Ha.
\newblock Stargan v2: Diverse image synthesis for multiple domains.
\newblock In {\em Proceedings of the IEEE/CVF conference on computer vision and
  pattern recognition}, 2020.

\bibitem{choudhury2021unsupervised}
Subhabrata Choudhury, Iro Laina, Christian Rupprecht, and Andrea Vedaldi.
\newblock Unsupervised part discovery from contrastive reconstruction.
\newblock {\em Advances in Neural Information Processing Systems}, 2021.

\bibitem{facciolo2017automatic}
Gabriele Facciolo, Carlo De~Franchis, and Enric Meinhardt-Llopis.
\newblock Automatic 3d reconstruction from multi-date satellite images.
\newblock In {\em Proceedings of the IEEE Conference on Computer Vision and
  Pattern Recognition Workshops}, pages 57--66, 2017.

\bibitem{fischer2014descriptor}
Philipp Fischer, Alexey Dosovitskiy, and Thomas Brox.
\newblock Descriptor matching with convolutional neural networks: a comparison
  to sift.
\newblock {\em arXiv preprint arXiv:1405.5769}, 2014.

\bibitem{frey1999estimating}
Brendan~J Frey and Nebojsa Jojic.
\newblock Estimating mixture models of images and inferring spatial
  transformations using the em algorithm.
\newblock In {\em Proceedings. 1999 IEEE Computer Society Conference on
  Computer Vision and Pattern Recognition (Cat. No PR00149)}. IEEE, 1999.

\bibitem{frey2003transformation}
Brendan~J. Frey and Nebojsa Jojic.
\newblock Transformation-invariant clustering using the em algorithm.
\newblock {\em IEEE Transactions on Pattern Analysis and Machine Intelligence},
  2003.

\bibitem{hamilton2022unsupervised}
Mark Hamilton, Zhoutong Zhang, Bharath Hariharan, Noah Snavely, and William~T
  Freeman.
\newblock Unsupervised semantic segmentation by distilling feature
  correspondences.
\newblock {\em arXiv preprint arXiv:2203.08414}, 2022.

\bibitem{huang2012learning}
Gary Huang, Marwan Mattar, Honglak Lee, and Erik Learned-Miller.
\newblock Learning to align from scratch.
\newblock {\em Advances in neural information processing systems}, 25, 2012.

\bibitem{huang2007unsupervised}
Gary~B Huang, Vidit Jain, and Erik Learned-Miller.
\newblock Unsupervised joint alignment of complex images.
\newblock In {\em 2007 IEEE 11th international conference on computer vision},
  pages 1--8. IEEE, 2007.

\bibitem{huber}
Peter~J. Huber.
\newblock {Robust Estimation of a Location Parameter}.
\newblock {\em The Annals of Mathematical Statistics}, 35(1):73 -- 101, 1964.

\bibitem{jaderberg2015spatial}
Max Jaderberg, Karen Simonyan, Andrew Zisserman, et~al.
\newblock Spatial transformer networks.
\newblock {\em Advances in neural information processing systems}, 28, 2015.

\bibitem{karras2020analyzing}
Tero Karras, Samuli Laine, Miika Aittala, Janne Hellsten, Jaakko Lehtinen, and
  Timo Aila.
\newblock Analyzing and improving the image quality of stylegan.
\newblock In {\em Proceedings of the IEEE/CVF conference on computer vision and
  pattern recognition}, pages 8110--8119, 2020.

\bibitem{Karras_2020_CVPR}
Tero Karras, Samuli Laine, Miika Aittala, Janne Hellsten, Jaakko Lehtinen, and
  Timo Aila.
\newblock Analyzing and improving the image quality of stylegan.
\newblock In {\em Proceedings of the IEEE/CVF Conference on Computer Vision and
  Pattern Recognition (CVPR)}, June 2020.

\bibitem{kasten2021layered}
Yoni Kasten, Dolev Ofri, Oliver Wang, and Tali Dekel.
\newblock Layered neural atlases for consistent video editing.
\newblock {\em ACM Transactions on Graphics (TOG)}, 40(6):1--12, 2021.

\bibitem{kemelmacher2012collection}
Ira Kemelmacher-Shlizerman and Steven~M Seitz.
\newblock Collection flow.
\newblock In {\em 2012 IEEE Conference on Computer Vision and Pattern
  Recognition}, pages 1792--1799. IEEE, 2012.

\bibitem{kingma2014adam}
Diederik~P Kingma and Jimmy Ba.
\newblock Adam: A method for stochastic optimization.
\newblock {\em arXiv preprint arXiv:1412.6980}, 2014.

\bibitem{learned2005data}
Erik~G Learned-Miller.
\newblock Data driven image models through continuous joint alignment.
\newblock {\em IEEE Transactions on Pattern Analysis and Machine Intelligence},
  28(2):236--250, 2005.

\bibitem{liu2009simultaneous}
Xiaoming Liu, Yan Tong, and Frederick~W Wheeler.
\newblock Simultaneous alignment and clustering for an image ensemble.
\newblock In {\em 2009 IEEE 12th International Conference on Computer Vision}.
  IEEE, 2009.

\bibitem{liu2020semantic}
Yanbin Liu, Linchao Zhu, Makoto Yamada, and Yi Yang.
\newblock Semantic correspondence as an optimal transport problem.
\newblock In {\em Proceedings of the IEEE/CVF Conference on Computer Vision and
  Pattern Recognition}, 2020.

\bibitem{lowe2004distinctive}
David~G Lowe.
\newblock Distinctive image features from scale-invariant keypoints.
\newblock {\em International journal of computer vision}, 60(2):91--110, 2004.

\bibitem{retiming}
Erika Lu, Forrester Cole, Tali Dekel, Weidi Xie, Andrew Zisserman, David
  Salesin, William~T. Freeman, and Michael Rubinstein.
\newblock Layered neural rendering for retiming people in video.
\newblock {\em {ACM} Trans. Graph.}, 2020.

\bibitem{omnimatte}
Erika Lu, Forrester Cole, Tali Dekel, Andrew Zisserman, William~T. Freeman, and
  Michael Rubinstein.
\newblock Omnimatte: Associating objects and their effects in video.
\newblock In {\em Proceedings of the IEEE/CVF Conference on Computer Vision and
  Pattern Recognition (CVPR)}, 2021.

\bibitem{mattar2012unsupervised}
Marwan~A Mattar, Allen~R Hanson, and Erik~G Learned-Miller.
\newblock Unsupervised joint alignment and clustering using bayesian
  nonparametrics.
\newblock {\em arXiv preprint arXiv:1210.4892}, 2012.

\bibitem{melas2022deep}
Luke Melas-Kyriazi, Christian Rupprecht, Iro Laina, and Andrea Vedaldi.
\newblock Deep spectral methods: A surprisingly strong baseline for
  unsupervised semantic segmentation and localization.
\newblock In {\em Proceedings of the IEEE/CVF Conference on Computer Vision and
  Pattern Recognition}, 2022.

\bibitem{miller2000learning}
Erik~G Miller, Nicholas~E Matsakis, and Paul~A Viola.
\newblock Learning from one example through shared densities on transforms.
\newblock In {\em Proceedings IEEE Conference on Computer Vision and Pattern
  Recognition. CVPR 2000 (Cat. No. PR00662)}. IEEE, 2000.

\bibitem{min2021convolutional}
Juhong Min and Minsu Cho.
\newblock Convolutional hough matching networks.
\newblock In {\em Proceedings of the IEEE/CVF Conference on Computer Vision and
  Pattern Recognition}, 2021.

\bibitem{min2019hyperpixel}
Juhong Min, Jongmin Lee, Jean Ponce, and Minsu Cho.
\newblock Hyperpixel flow: Semantic correspondence with multi-layer neural
  features.
\newblock In {\em Proceedings of the IEEE/CVF International Conference on
  Computer Vision}, 2019.

\bibitem{min2019spair}
Juhong Min, Jongmin Lee, Jean Ponce, and Minsu Cho.
\newblock Spair-71k: A large-scale benchmark for semantic correspondence.
\newblock {\em arXiv prepreint arXiv:1908.10543}, 2019.

\bibitem{min2020learning}
Juhong Min, Jongmin Lee, Jean Ponce, and Minsu Cho.
\newblock Learning to compose hypercolumns for visual correspondence.
\newblock In {\em European Conference on Computer Vision}. Springer, 2020.

\bibitem{mobahi2014compositional}
Hossein Mobahi, Ce Liu, and William~T Freeman.
\newblock A compositional model for low-dimensional image set representation.
\newblock In {\em Proceedings of the IEEE Conference on Computer Vision and
  Pattern Recognition}, pages 1322--1329, 2014.

\bibitem{mokady2022self}
Ron Mokady, Omer Tov, Michal Yarom, Oran Lang, Inbar Mosseri, Tali Dekel,
  Daniel Cohen-Or, and Michal Irani.
\newblock Self-distilled stylegan: Towards generation from internet photos.
\newblock In {\em ACM SIGGRAPH 2022 Conference Proceedings}, pages 1--9, 2022.

\bibitem{monnier2020deep}
Tom Monnier, Thibault Groueix, and Mathieu Aubry.
\newblock Deep transformation-invariant clustering.
\newblock {\em Advances in Neural Information Processing Systems}, 2020.

\bibitem{peebles2022gan}
William Peebles, Jun-Yan Zhu, Richard Zhang, Antonio Torralba, Alexei~A Efros,
  and Eli Shechtman.
\newblock Gan-supervised dense visual alignment.
\newblock In {\em Proceedings of the IEEE/CVF Conference on Computer Vision and
  Pattern Recognition}, pages 13470--13481, 2022.

\bibitem{peng2012rasl}
Yigang Peng, Arvind Ganesh, John Wright, Wenli Xu, and Yi Ma.
\newblock Rasl: Robust alignment by sparse and low-rank decomposition for
  linearly correlated images.
\newblock {\em IEEE transactions on pattern analysis and machine intelligence},
  34(11):2233--2246, 2012.

\bibitem{pixabay}
Pixabay.
\newblock https://pixabay.com/.

\bibitem{rocco2017convolutional}
Ignacio Rocco, Relja Arandjelovic, and Josef Sivic.
\newblock Convolutional neural network architecture for geometric matching.
\newblock In {\em Proceedings of the IEEE conference on computer vision and
  pattern recognition}, 2017.

\bibitem{rocco2018end}
Ignacio Rocco, Relja Arandjelovi{\'c}, and Josef Sivic.
\newblock End-to-end weakly-supervised semantic alignment.
\newblock In {\em Proceedings of the IEEE Conference on Computer Vision and
  Pattern Recognition}, pages 6917--6925, 2018.

\bibitem{rocco2018neighbourhood}
Ignacio Rocco, Mircea Cimpoi, Relja Arandjelovi{\'c}, Akihiko Torii, Tomas
  Pajdla, and Josef Sivic.
\newblock Neighbourhood consensus networks.
\newblock {\em Advances in neural information processing systems}, 2018.

\bibitem{rubinstein2013unsupervised}
Michael Rubinstein, Armand Joulin, Johannes Kopf, and Ce Liu.
\newblock Unsupervised joint object discovery and segmentation in internet
  images.
\newblock In {\em Proceedings of the IEEE conference on computer vision and
  pattern recognition}, pages 1939--1946, 2013.

\bibitem{seo2018attentive}
Paul~Hongsuck Seo, Jongmin Lee, Deunsol Jung, Bohyung Han, and Minsu Cho.
\newblock Attentive semantic alignment with offset-aware correlation kernels.
\newblock In {\em Proceedings of the European Conference on Computer Vision
  (ECCV)}, 2018.

\bibitem{shutterstock}
Shutterstock.
\newblock https://www.shutterstock.com/.

\bibitem{simeoni2021localizing}
Oriane Sim{\'e}oni, Gilles Puy, Huy~V Vo, Simon Roburin, Spyros Gidaris, Andrei
  Bursuc, Patrick P{\'e}rez, Renaud Marlet, and Jean Ponce.
\newblock Localizing objects with self-supervised transformers and no labels.
\newblock {\em arXiv preprint arXiv:2109.14279}, 2021.

\bibitem{simo2015discriminative}
Edgar Simo-Serra, Eduard Trulls, Luis Ferraz, Iasonas Kokkinos, Pascal Fua, and
  Francesc Moreno-Noguer.
\newblock Discriminative learning of deep convolutional feature point
  descriptors.
\newblock In {\em Proceedings of the IEEE international conference on computer
  vision}, pages 118--126, 2015.

\bibitem{Tang_2021_ICCV}
Xiao Tang, Tianyu Wang, and Chi-Wing Fu.
\newblock Towards accurate alignment in real-time 3d hand-mesh reconstruction.
\newblock In {\em Proceedings of the IEEE/CVF International Conference on
  Computer Vision (ICCV)}, pages 11698--11707, October 2021.

\bibitem{tola2009daisy}
Engin Tola, Vincent Lepetit, and Pascal Fua.
\newblock Daisy: An efficient dense descriptor applied to wide-baseline stereo.
\newblock {\em IEEE transactions on pattern analysis and machine intelligence},
  32(5):815--830, 2009.

\bibitem{tumanyan2022splicing}
Narek Tumanyan, Omer Bar-Tal, Shai Bagon, and Tali Dekel.
\newblock Splicing vit features for semantic appearance transfer.
\newblock In {\em Proceedings of the IEEE/CVF Conference on Computer Vision and
  Pattern Recognition}, pages 10748--10757, 2022.

\bibitem{vaze2022generalized}
Sagar Vaze, Kai Han, Andrea Vedaldi, and Andrew Zisserman.
\newblock Generalized category discovery.
\newblock In {\em Proceedings of the IEEE/CVF Conference on Computer Vision and
  Pattern Recognition}, 2022.

\bibitem{vedaldi2008joint}
Andrea Vedaldi, Gregorio Guidi, and Stefano Soatto.
\newblock Joint data alignment up to (lossy) transformations.
\newblock In {\em 2008 IEEE Conference on Computer Vision and Pattern
  Recognition}. IEEE, 2008.

\bibitem{WahCUB_200_2011}
C. Wah, S. Branson, P. Welinder, P. Perona, and S. Belongie.
\newblock .
\newblock Technical Report CNS-TR-2011-001, California Institute of Technology,
  2011.

\bibitem{wang2022self}
Yangtao Wang, Xi Shen, Shell~Xu Hu, Yuan Yuan, James~L Crowley, and Dominique
  Vaufreydaz.
\newblock Self-supervised transformers for unsupervised object discovery using
  normalized cut.
\newblock In {\em Proceedings of the IEEE/CVF Conference on Computer Vision and
  Pattern Recognition}, 2022.

\bibitem{yu2015lsun}
Fisher Yu, Ari Seff, Yinda Zhang, Shuran Song, Thomas Funkhouser, and Jianxiong
  Xiao.
\newblock Lsun: Construction of a large-scale image dataset using deep learning
  with humans in the loop.
\newblock {\em arXiv preprint arXiv:1506.03365}, 2015.

\bibitem{zhou2015flowweb}
Tinghui Zhou, Yong Jae~Lee, Stella~X Yu, and Alyosha~A Efros.
\newblock Flowweb: Joint image set alignment by weaving consistent, pixel-wise
  correspondences.
\newblock In {\em Proceedings of the IEEE Conference on Computer Vision and
  Pattern Recognition}, pages 1191--1200, 2015.

\bibitem{zhu2014averageexplorer}
Jun-Yan Zhu, Yong~Jae Lee, and Alexei~A Efros.
\newblock Averageexplorer: Interactive exploration and alignment of visual data
  collections.
\newblock {\em ACM Transactions on Graphics (TOG)}, 33(4):1--11, 2014.

\end{thebibliography}
}

\clearpage
\appendix
\section{Implementation Details}\label{app:impl-details}

\subsection{General}
We use images of size $256\times 256$ (using border padding for non-square image), which is the resolution the STN takes as input. The rigid STN resizes the images to the resolution of the atlas, which is set to be $W_{\mathcal{A}}=H_{\mathcal{A}}=128$. For feature extraction, we use \texttt{dino\_vits8} ($D = 384$) with stride 4, as in \cite{amir2021deep}. We extract the keys from the original images and bilinearly upsample them to the atlas resolution.

\subsection{Spatial Transformer Architecture Details}\label{app:stn-arch}
We use the same architectures for the STNs as in \cite{peebles2022gan}. Both architectures are based on the design of the ResNet-based discriminator from StyleGAN2~\cite{Karras_2020_CVPR}. Tables~\ref{tab:rigid-stn} and \ref{tab:non-rigid-stn} detail the layers of the rigid and non-rigid STN respectively, and \ref{tab:resblock} and \ref{tab:convLblock} detail the building blocks.

\paragraph{Rigid STN.} 
The rigid mapping network consists of a ResNet backbone with a fully-connected layer at the end, which outputs four logits $(o_1, o_2, o_3, o_4)$, to which the following activations are applied to obtain the transformation parametrization: 
\begin{align}
\theta = \pi\cdot \tanh(o_1) \quad,\quad R = \begin{bmatrix}\cos\theta & -\sin\theta \\ \sin\theta & \cos\theta\end{bmatrix} \\
s = \exp(o_2) \\
\vv{t} = \begin{bmatrix}o_3 \\ o_4\end{bmatrix}
\end{align}

\paragraph{Non-Rigid STN.} The non rigid mapping network consists of a ResNet backbone that outputs a $16\times 16$ feature grid which is then fed to two small convolutional networks: the first outputs a $16\times 16$ coarse flow field, and the second outputs weights which are used to perform $\times 8$ upsampling of the coarse flow field to the size of $128\times 128 = H_\mathcal{A}\times W_\mathcal{A}$. The final flow is bilinearly upsampled in case of applying backward warp on inputs of resolution higher than $128$.

When composing both networks, the affine matrix given by the rigid STN is applied to the non-rigid flow which results in the final sampling grid used to congeal the original input image, DINO-ViT features and saliency mask.

\begin{table}[b!]
  \centering
  \scriptsize
  \begin{tabular*}{0.5\textwidth}{@{}lccc@{}}
    \toprule
    block & layer  & output size \\
    \midrule
    0  & bilinear downsample (using conv2d) &  $3\times 128\times 128$ \\
    1  & convL(64, 1, 1, 0, fusedLeakyReLU) &  $64\times 128\times 128$ \\
    2  & ResBlock((64, 3, 1, 1), (128, 3, 2, 0), (128, 1, 2, 0)) & $128\times 64\times 64$ \\
    3  & ResBlock((128, 3, 1, 1), (512, 3, 2, 0), (512, 1, 2, 0)) & $512\times 32\times 32$ \\
    4  & ResBlock((512, 3, 1, 1), (512, 3, 2, 0), (512, 1, 2, 0)) & $512\times 16\times 16$ \\
    5  & ResBlock((512, 3, 1, 1), (512, 3, 2, 0), (512, 1, 2, 0)) & $512\times 8\times 8$ \\
    6  & ResBlock((512, 3, 1, 1), (512, 3, 2, 0), (512, 1, 2, 0)) &  $512\times 4\times 4$\\
    7  & convL(512, 1, 2, 0, fusedLeakyReLU) & $512\times 4\times 4$ (flattened)\\    
    8  & linear + fusedLeakyReLU &   $1\times 512$\\
    9  & linear &  $1\times 4$\\

    \bottomrule
  \end{tabular*}
  \caption{\emph{Architecture of rigid STN.}}
  \label{tab:rigid-stn}
\end{table}

\begin{table}
  \centering
  \scriptsize
  \begin{tabular*}{0.5\textwidth}{@{}lccc@{}}
    \toprule
    block & layer  & output size \\
    \midrule
    0  & convL(64, 1, 1, 0, fusedLeakyReLU) &  $64\times 128\times 128$ \\
    1  & ResBlock((64, 3, 1, 1), (128, 3, 2, 0), (128, 1, 2, 0)) & $128\times 64\times 64$ \\
    2  & ResBlock((128, 3, 1, 1), (512, 3, 2, 0), (512, 1, 2, 0)) & $512\times 32\times 32$ \\
    3  & ResBlock((512, 3, 1, 1), (512, 3, 2, 0), (512, 1, 2, 0)) & $512\times 16\times 16$ \\
    4  & ResBlock((512, 3, 1, 1), (512, 3, 1, 1), (512, 1, 1, 0)) & $512\times 16\times 16$ \\
    5  & convL(512, 3, 1, 1, fusedLeakyReLU) & $512\times 16\times 16$\\
    6  & conv2d(512, 3, 1, 1) + ReLU + conv2d(2, 3, 1, 1) &   \makecell{$2\times 16 \times 16$ \\(coarse flow)}\\
    7 & conv2d(512, 3, 1, 1) + ReLU + conv2d(576, 3, 1, 1) &  \makecell{$576\times 16 \times 16$ \\(upsampling weights)}\\
    \bottomrule
  \end{tabular*}
  \caption{\emph{Architecture of non-rigid STN.}}
  \label{tab:non-rigid-stn}
\end{table}

\begin{table}
\scriptsize 
  \centering
  \begin{tabular*}{0.5\textwidth}{@{}lccc@{}}
    \toprule
    \multicolumn{3}{c}{\makecell{ResBlock((channels1, kernel1, stride1, padding1), \\(channels2, kernel2, stride2, padding2), \\(channels3, kernel3, stride3, padding3))}} \\
    \midrule
    order & type & layer \\
    \midrule
    0 & conv1 & convL(channels1, kernel1, stride1, padding1, fusedLeakyReLU) \\
    1 & conv2 & convL(channels2, kernel2, stride2, padding2, blur, fusedLeakyReLU) \\
    2 & skip & convL(channels3, kernel3, stride3, padding3, blur)\\
    \bottomrule
  \end{tabular*}
  \caption{\emph{Architecture of a ResBlock.}}
  \label{tab:resblock}
\end{table}

\begin{table}
  \centering
  \footnotesize
  \begin{tabular}{@{}lcc@{}}
    \toprule
    \multicolumn{2}{c}{convL(channels, kernel, stride, padding, blur, fusedLeakyReLU)} \\
    \midrule
    order & layer \\
    \midrule
    0 & blur (upfirdn2d) \\
    1 & conv2d(channels, kernel, stride, padding) \\
    2 & fusedLeakyReLU\\
    \bottomrule
  \end{tabular}
  \caption{\emph{Architecture of a convL layer.}}
  \label{tab:convLblock}
\end{table}

\subsection{Loss Terms}\label{app:loss-terms}

All losses except for $\mathcal{L}_{sparsity}$ and $\mathcal{L}_{scale}$ are applied in the atlas space within the boundaries of the backward warped images. 
Formally, for $\mathcal{L}_{keys}$, $ \mathcal{L}_{saliency}$, $\mathcal{L}_{mag}$, $ \mathcal{L}_{smooth}$ and $\mathcal{L}_{center}$ the sum in the atlas space is taken over the indices $\{\vv{x}_{\mathcal{A}} \,\,|\,\, \mathcal{M}(I_i,\vv{x}_\mathcal{A}) \in I_i \}$.

We detach the atlas saliency $S_\mathcal{A}$ for losses which should not have an impact on the joint saliency, which are $\mathcal{L}_{keys}$ and local $\mathcal{L}_{smooth}$ (see details next).

\paragraph{Rigidity Loss $\mathcal{L}_{smooth}$.}
Recall the term $\mathcal{L}_{smooth}$, defined as in \cite{kasten2021layered}, is formally defined by:
 \begin{equation}
\mathcal{L}_{smooth} = \frac{1}{N\cdot N_\mathcal{A}}\sum_{i=1}^N \sum_{x_\mathcal{A}}  \left(\norm{J^TJ}_F+\norm{\left(J^TJ\right)^{-1}}_F\right)
\end{equation}
where $N$ is the number of images, $N_\mathcal{A}$ is the number of pixels in the atlas, and $J$ is the Jacobian matrix of $\mathcal{M}$ at $\vv{x}_\mathcal{A}$. The term is used to prevent the non-rigid mapping from distorting the shared content by encouraging as rigid as possible mapping. The Jacobian matrix is defined by:

\begin{equation}
J = \frac{1}{\Delta} \begin{bmatrix} \vv{j}_1 & \vv{j}_2  \end{bmatrix} \in \mathbb{R}^{2\times 2 }
\end{equation}
where 
\begin{align}
\vv{j}_1=\mathcal{M}\left(I_i, \vv{x}_\mathcal{A}+\Delta \cdot \begin{bmatrix}1 \\0\end{bmatrix}\right) -  \mathcal{M}(I_i, \vv{x}_\mathcal{A}), \\
\vv{j}_2=\mathcal{M}\left(I_i, \vv{x}_\mathcal{A}+\Delta \cdot \begin{bmatrix}0 \\1\end{bmatrix}\right) -  \mathcal{M}(I_i, \vv{x}_\mathcal{A})
\end{align}
and $\Delta$ corresponds to the offset in pixels. This encourages both singular values of $J$ to be 1, which is what is required for a rigid mapping. We apply both local and global constraints, with $\Delta=1$ and $\Delta=20$ respectively. 

In practice, similarly to the other losses, we apply the local rigidity loss only within the salient parts of the atlas (similarly to Eq.~\eqref{eq:keys-loss}).

\subsection{Training}
We train the atlas and the STNs jointly. We first bootstrap the rigid STN for 1000 epochs, and then train both the rigid and non-rigid components on their own (separate objective function), when only the non-rigid network affects the atlas training. We train for a total of 8000 epochs, and use Adam optimizer \cite{kingma2014adam} with a learning rate of $1\cdot 10^{-4}$ for the STNs and $8\cdot 10^{-4}$ for the atlas. Training on a set of 10 images on a Tesla V100-SXM2-32GB takes 1.2 hrs and uses 3.4GB of GPU memory, or 1.8 hrs and 5.3GB of GPU memory in case of including horizontal flips.

\paragraph{Loss coefficients.} The loss coefficients we used for all experiments are as follows: 

\begin{equation}
\mathcal{L} = \mathcal{L}_{keys} + \lambda_s\mathcal{L}_{saliency} +  \lambda_r\mathcal{L}_{reg_{\mathcal{M}}} + \lambda_a\mathcal{L}_{reg_\mathcal{A}} 
\label{eq:objective1}
\end{equation}
with $\lambda_s = 1.25, \lambda_r = 0.025, \lambda_a = 0.75$.

\begin{equation}
\mathcal{L}_{keys} = \lambda_l L_2 + \mathcal{D}_{cos}
\end{equation}
with $ \lambda_l = 0.875$.

\begin{equation}
\mathcal{L}_{reg_\mathcal{M}} = \lambda_{s_1} \mathcal{L}_{scale} + \lambda_{s_2}\mathcal{L}_{mag} + \mathcal{L}_{smooth}
\end{equation}
with $\lambda_{s_1} = 8, \lambda_{s_2} = 80$. As mentioned in Appendix~\ref{app:loss-terms}, we apply $\mathcal{L}_{smooth}$ both locally and globally, namely:
\begin{equation}
\mathcal{L}_{smooth} = \mathcal{L}_{smooth}^{\Delta=1} + \lambda_{s_3}\mathcal{L}_{smooth}^{\Delta=20}
\end{equation}
with $\lambda_{s_3} = 3.5$.

\begin{equation}
\mathcal{L}_{reg_\mathcal{A}} = \mathcal{L}_{center} + \lambda_p\mathcal{L}_{sparsity}
\end{equation}
with $\lambda_p = 0.075$. As mentioned in Sec.~\ref{sec:training}, we apply $\lambda_p\mathcal{L}_{sparsity}$ both to the atlas saliency and the atlas keys, namely:

\begin{equation}
\mathcal{L}_{sparsity} =  \mathcal{L}_{sparsity}^{S_\mathcal{A}} + \lambda_{k}\mathcal{L}_{sparsity}^{K_\mathcal{A}}
\end{equation}
with $\lambda_{k} = 0.044$. For $\mathcal{L}_{sparsity}^{S_\mathcal{A}}$ we set the relative weight between the L1- and L0-approximation terms to be $\gamma=2$.

The entire objective function is multiplied by a scalar $c\cdot \mathcal{L}$, where $c=4000$.

\subsubsection{Congealing under extreme deformations.}\label{app:ext-d} Similarly to \cite{peebles2022gan},  we include an option of allowing horizontal flips in a given set, which is also used for the training of subsets of SPair 71K and CUB-200-2011 (Sec.~\ref{sec:quantitative}). The flipping is done during training: we train the STNs with both the original images and the flipped images, and update the atlas only according to the orientation that currently has a lower semantic loss (keys loss). 

Due to the extreme deformations present in the subsets of SPair 71K and CUB-200-2011, to increase robustness, we further reduce the local and global rigidity coefficients to be $\times 0.25$ its original value and the global rigidity to be $\lambda_{s_3}=0.9$. In addition,  the atlas representation is gradually updated during training, i.e., the images used to update the atlas are added one-by-one where every 100 epochs the image with the lowest key loss is added. We observed that this training scheme is more stable and allows faster convergence for these sets. For the \emph{Bicycle} set of SPair-71K, since many images contain only one wheel, 
we fix the atlas with the image that is most semantically similar to the average keys of the set and train the set with a fixed atlas.

\begin{figure*}[t!]
    \centering
    \includegraphics[width=\textwidth]{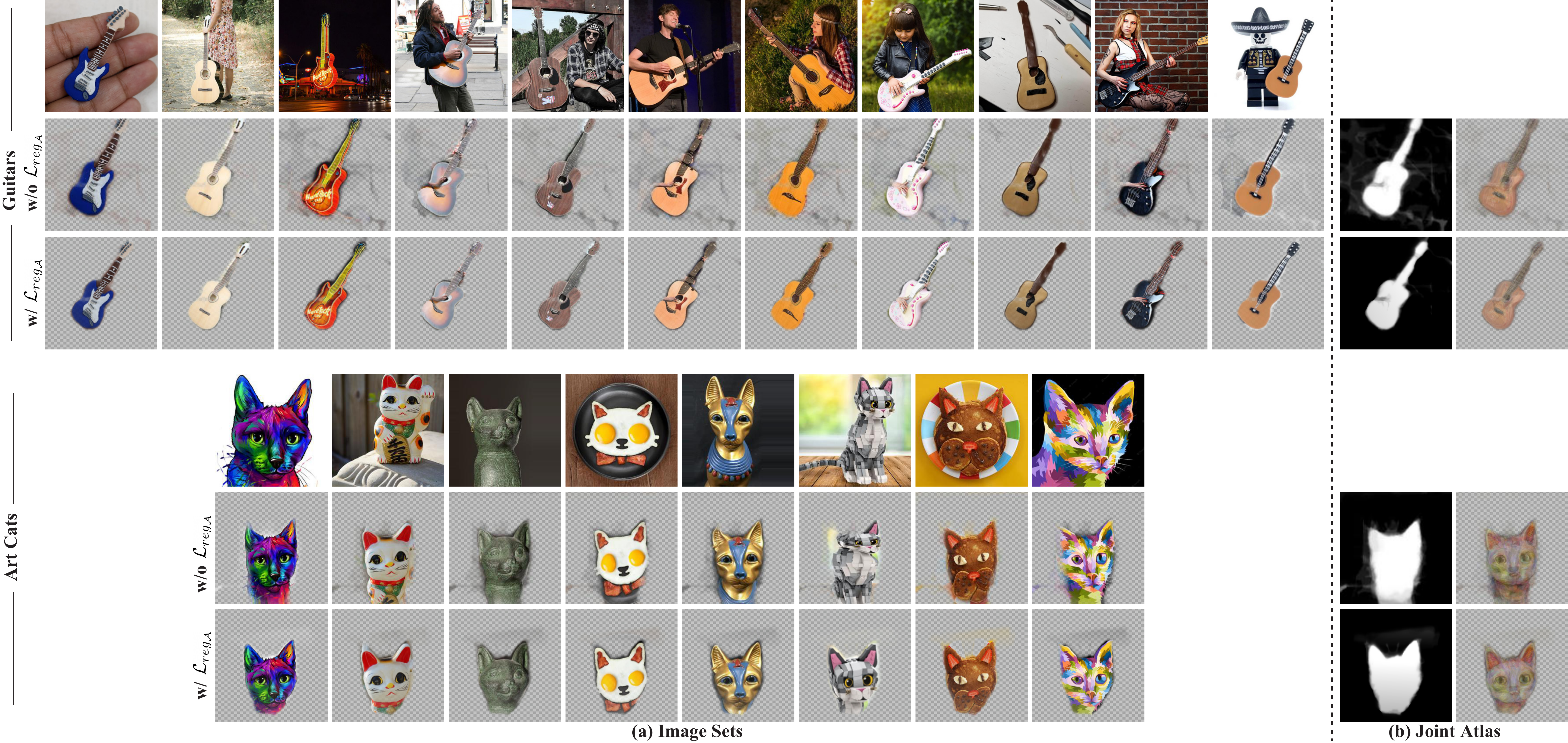}
    \caption{\emph{Results without ${\bf \mathcal{L}_{reg_\mathcal{A}}}$.}  For each set, we show the original images, and the congealed images with and without ${\bf \mathcal{L}_{reg_\mathcal{A}}}$; on the right are the average image in atlas space and the atlas saliency. This regularization encourages cleaner atlases,  e.g., ignoring background clutter in \emph{Guitars}, and allows us to better capture the dominant shared content, e.g., focusing only on the face in \emph{Art Cats} allows us to better align the third cat from the right.
    }
    \label{fig:noRegA}
\end{figure*}

\section{Point Correspondence Between A Pair of Images}\label{app:correspondences}
As in~\cite{peebles2022gan}, our method can find dense correspondences between a pair of images. For each image pair $\{I_A, I_B\}$, we transfer the ground truth keypoints  $\mathbf{k}_A \in I_A$ to $I_B$. This is done by mapping $\bf{k}_A$ to the atlas, obtaining $\mathbf{k}_\mathcal{A}$ then mapping it to $I_B$. Recall that our mapping $\mathcal{M}=\mathcal{M}_r \circ \mathcal{M}_f$ is defined from the atlas to each image. For mapping $\mathbf{k}_A$ to the atlas, we first compute the inverse of the rigid transformation, which has a closed-form solution (inverse of an affine matrix). Then, since there is no closed form for obtaining the inverse of $\mathcal{M}_f$, we follow ~\cite{peebles2022gan}, and approximate the inverse using nearest neighbors. Finally, we map $\mathbf{k}_\mathcal{A}$ to $I_B$ by bilinearly sampling the mapping grid of $I_B$.

\section{Ablation Study:  No Atlas Regularization}\label{app:no-reg-a} 
As discussed in the main paper (Sec.~\ref{sec:ablation}), the sparsity regularization on the atlas assists our framework in capturing the dominant shared content, while ignoring noise and background clutter. 

Sample cases can be seen in Fig.~\ref{fig:noRegA}: for \emph{Guitars}, some background content is captured by the atlas w/o this regularization. In \emph{Art Cats}, the sparsity regularization allows us to only focus on the face, while ignoring unshared regions even if they are initially considered to be salient (cat's body, third column from the right).

\end{document}